\documentclass[journal]{IEEEtran}
\usepackage{cite}

\ifCLASSINFOpdf
  \usepackage[pdftex]{graphicx}
\else
\fi

\usepackage{url}
\usepackage{pifont}
\usepackage{graphicx}
\usepackage{tablefootnote}
\usepackage{soul, color}
\usepackage{dblfloatfix}

\usepackage{scalerel}
\usepackage{tikz}
\usetikzlibrary{svg.path}

\definecolor{orcidlogocol}{HTML}{A6CE39}
\tikzset{
  orcidlogo/.pic={
    \fill[orcidlogocol] svg{M256,128c0,70.7-57.3,128-128,128C57.3,256,0,198.7,0,128C0,57.3,57.3,0,128,0C198.7,0,256,57.3,256,128z};
    \fill[white] svg{M86.3,186.2H70.9V79.1h15.4v48.4V186.2z}
                 svg{M108.9,79.1h41.6c39.6,0,57,28.3,57,53.6c0,27.5-21.5,53.6-56.8,53.6h-41.8V79.1z M124.3,172.4h24.5c34.9,0,42.9-26.5,42.9-39.7c0-21.5-13.7-39.7-43.7-39.7h-23.7V172.4z}
                 svg{M88.7,56.8c0,5.5-4.5,10.1-10.1,10.1c-5.6,0-10.1-4.6-10.1-10.1c0-5.6,4.5-10.1,10.1-10.1C84.2,46.7,88.7,51.3,88.7,56.8z};
  }
}

\newcommand\orcidicon[1]{\href{https://orcid.org/#1}{\mbox{\scalerel*{
\begin{tikzpicture}[yscale=-1,transform shape]
\pic{orcidlogo};
\end{tikzpicture}
}{|}}}}

\usepackage{hyperref}

\begin{document}
\title{Video Summarization Using Deep Neural Networks: A Survey}

\author{Evlampios~Apostolidis~\orcidicon{0000-0001-5376-7158},
Eleni~Adamantidou~\orcidicon{0000-0002-1981-6902},
Alexandros I.~Metsai~\orcidicon{0000-0002-6978-6924},
Vasileios~Mezaris~\orcidicon{0000-0002-0121-4364},~\IEEEmembership{Senior~Member,~IEEE,}
and~Ioannis~Patras~\orcidicon{0000-0003-3913-4738},~\IEEEmembership{Senior~Member,~IEEE}
\thanks{This work was supported by the EU Horizon 2020 research and innovation programme under grant agreements H2020-780656 ReTV and H2020-951911 AI4Media, and by EPSRC under grant No. EP/R026424/1.}
\thanks{E. Apostolidis is with the Information Technologies Institute / Centre for Research and Technology Hellas, GR-57001, Thermi, Thessaloniki, Greece, and with the Queen Mary University of London, Mile End Campus, E14NS, London, UK, e-mail: apostolid@iti.gr.}
\thanks{E. Adamantidou, A. I. Metsai and V. Mezaris are with the Information Technologies Institute / Centre for Research and Technology Hellas, GR-57001, Thermi, Thessaloniki, Greece, e-mail: \{adamelen, alexmetsai, bmezaris\}@iti.gr.}
\thanks{I. Patras is with the Queen Mary University of London, Mile End Campus, E14NS, London, UK, e-mail: i.patras@qmul.ac.uk.}}

\markboth{Accepted version}%
{Apostolidis \MakeLowercase{\textit{et al.}}: Video Summarization Using Deep Neural Networks: A Survey}

\maketitle

\begin{abstract}
Video summarization technologies aim to create a concise and complete synopsis by selecting the most informative parts of the video content. Several approaches have been developed over the last couple of decades and the current state of the art is represented by methods that rely on modern deep neural network architectures. This work focuses on the recent advances in the area and provides a comprehensive survey of the existing deep-learning-based methods for generic video summarization. After presenting the motivation behind the development of technologies for video summarization, we formulate the video summarization task and discuss the main characteristics of a typical deep-learning-based analysis pipeline. Then, we suggest a taxonomy of the existing algorithms and provide a systematic review of the relevant literature that shows the evolution of the deep-learning-based video summarization technologies and leads to suggestions for future developments. We then report on protocols for the objective evaluation of video summarization algorithms and we compare the performance of several deep-learning-based approaches. Based on the outcomes of these comparisons, as well as some documented considerations about the amount of annotated data and the suitability of evaluation protocols, we indicate potential future research directions.
\end{abstract}


\IEEEpeerreviewmaketitle

\section{Introduction}
In July 2015, YouTube revealed that it receives over 400 hours of video content every single minute, which translates to 65.7 years' worth of content uploaded every day\footnote{https://www.tubefilter.com/2015/07/26/youtube-400-hours-content-every-minute/}. Since then, we are experiencing an even stronger engagement of consumers with both online video platforms and devices (e.g., smart-phones, wearables etc.) that carry powerful video recording sensors and allow instant uploading of the captured video on the Web. According to newer estimates, YouTube now receives 500 hours of video per minute\footnote{https://blog.youtube/press/}; and YouTube is just one of the many video hosting platforms (e.g., DailyMotion, Vimeo), social networks (e.g., Facebook, Twitter, Instagram), and online repositories of media and news organizations that host large volumes of video content. So, how is it possible for someone to efficiently navigate within endless collections of videos, and find the video content that s/he is looking for? The answer to this question comes not only from video retrieval technologies but also from technologies for automatic video summarization. The latter allow generating a concise synopsis that conveys the important parts of the full-length video. Given the plethora of video content on the Web, effective video summarization facilitates viewers' browsing of and navigation in large video collections, thus increasing viewers' engagement and content consumption. 

The application domain of automatic video summarization is wide and includes (but is not limited to) the use of such technologies by media organizations (after integrating such techniques into their content management systems), to allow effective indexing, browsing, retrieval and promotion of their media assets; and video sharing platforms, to improve viewing experience, enhance viewers' engagement and increase content consumption. In addition, video summarization that is tailored to the requirements of particular content presentation scenarios can be used for e.g., generating trailers or teasers of movies and episodes of a TV series; presenting the highlights of an event (e.g., a sports game, a music band performance, or a public debate); and creating a video synopsis with the main activities that took place over e.g., the last 24hrs of recordings of a surveillance camera, for time-efficient progress monitoring or security purposes.

A number of surveys on video summarization have already appeared in the literature. In one of the first works, Barbieri et al. (2003) \cite{10.1117/12.515733} classify the relevant bibliography according to several aspects of the summarization process, namely the targeted scenario, the type of visual content, and the characteristics of the summarization approach. In another early study, Li et al. (2006) \cite{1621451} divide the existing summarization approaches into utility-based methods that use attention models to identify the salient objects and scenes, and structure-based methods that build on the video shots and scenes. Truong et al. (2007) \cite{10.1145/1198302.1198305} discuss a variety of attributes that affect the outcome of a summarization process, such as the video domain, the granularity of the employed video fragments, the utilized summarization methodology, and the targeted type of summary. Money et al. (2008) \cite{MONEY2008121} divide the bibliography into methods that rely on the analysis of the video stream, methods that process contextual video metadata, and hybrid approaches that rely on both types of the aforementioned data. Jiang et al. (2009) \cite{Jiang2009} discuss a few characteristic video summarization approaches, that include the extraction of low-level visual features for assessing frame similarity or performing clustering-based key-frame selection; the detection of the main events of the video using motion descriptors; and the identification of the video structure using Eigen-features. Hu et al. (2011) \cite{5729374} classify the summarization methods into those that target minimum visual redundancy, those that rely on object or event detection, and others that are based on multimodal integration. Ajmal et al. (2012) \cite{10.1007/978-3-642-33564-8_1} similarly classify the relevant literature in clustering-based methods, approaches that rely on detecting the main events of the story, etc. Nevertheless, all the aforementioned works (published between 2003 and 2012) report on early approaches to video summarization; they do not present how the summarization landscape has evolved over the last years and especially after the introduction of deep learning algorithms.

The more recent study of Molino et al. (2017) \cite{7750564} focuses on egocentric video summarization and discusses the specifications and the challenges of this task. In another recent work, Basavarajaiah et al. (2019) \cite{10.1145/3355398} provide a classification of various summarization approaches, including some recently-proposed deep-learning-based methods; however, their work mainly focuses on summarization algorithms that are directly applicable on the compressed domain. Finally, the survey of Vivekraj et al. (2019) \cite{10.1145/3347712} presents the relevant bibliography based on a two-way categorization, that relates to the utilized data modalities during the analysis and the incorporation of human aspects. With respect to the latter, it further splits the relevant literature into methods that create summaries by modeling the human understanding and preferences (e.g., using attention models, the semantics of the visual content, or ground-truth annotations and machine-learning algorithms), and conventional approaches that rely on the statistical processing of low-level features of the video. Nevertheless, none of the above surveys presents in a comprehensive manner the current developments towards generic video summarization, that are tightly related to the growing use of advanced deep neural network architectures for learning the summarization task. As a matter of fact, the relevant research area is a very active one as several new approaches are being presented every year in highly-ranked peer-reviewed journals and international conferences. In this survey, we study in detail more than $40$ different deep-learning-based video summarization algorithms among the relevant works that have been proposed over the last five years. In addition, a comparison of the summarization performance reported in the most recent deep-learning-based methods against the performance reported in other more conventional approaches, e.g., \cite{10.1145/3347712,7298928,Zhang2016SummaryTE,9293322}, shows that in most cases the deep-learning-based methods significantly outperform more traditional approaches that rely on weighted fusion, sparse subset selection or data clustering algorithms, and represent the current state of the art in automatic video summarization. Motivated by these observations, we aim to fill this gap in the literature by presenting the relevant bibliography on deep-learning-based video summarization and also discussing other aspects that are associated with it, such as the protocols used for evaluating video summarization.

This article begins, in Section \ref{problem_statement}, by defining the problem of automatic video summarization and presenting the most prominent types of video summary. Then, it provides a high-level description of the analysis pipeline of deep-learning-based video summarization algorithms, and introduces a taxonomy of the relevant literature according to the utilized data modalities, the adopted training strategy, and the implemented learning approaches. Finally, it discusses aspects that relate to the generated summary, such as the desired properties of a static (frame-based) video summary and the length of a dynamic (fragment-based) video summary. Section \ref{SoA} builds on the introduced taxonomy to systematically review the relevant bibliography. A primary categorization is made according to the use or not of human-generated ground-truth data for learning, and a secondary categorization is made based on the adopted learning objective or the utilized data modalities by each different class of methods. For each one of the defined classes, we illustrate the main processing pipeline and report on the specifications of the associated summarization algorithms. After presenting the relevant bibliography, we provide some general remarks that reflect how the field has evolved, especially over the last five years, highlighting the pros and cons of each class of methods. Section \ref{evaluation} continues with an in-depth discussion on the utilized datasets and the different evaluation protocols of the literature. Following, Section \ref{sec:perf_comp} discusses the findings of extensive performance comparisons that are based on the results reported in the relevant papers, indicates the most competitive methods in the fields of (weakly-)supervised and unsupervised video summarization, and examines whether there is a performance gap between these two main types of approaches. Based on the surveyed bibliography, in Section \ref{future_directions} we propose potential future directions to further advance the current state of the art in video summarization. Finally, Section \ref{sec:conclusions} concludes this work by briefly outlining the core findings of the conducted study.

\section{Problem statement}
\label{problem_statement}

Video summarization aims to generate a short synopsis that summarizes the video content by selecting its most informative and important parts. The produced summary is usually composed of a set of representative video frames (a.k.a. video key-frames), or video fragments (a.k.a. video key-fragments) that have been stitched in chronological order to form a shorter video. The former type of a video summary is known as video storyboard, and the latter type is known as video skim. One advantage of video skims over static sets of frames is the ability to include audio and motion elements that offer a more natural story narration and potentially enhance the expressiveness and the amount of information conveyed by the video summary. Furthermore, it is often more entertaining and interesting for the viewer to watch a skim rather than a slide show of frames~\cite{Li_HP_TR}. On the other hand, storyboards are not restricted by timing or synchronization issues and, therefore, they offer more flexibility in terms of data organization for browsing and navigation purposes \cite{4265619,4284941}.

\begin{figure*}[t]
\begin{center}
   \includegraphics[width=0.98\textwidth]{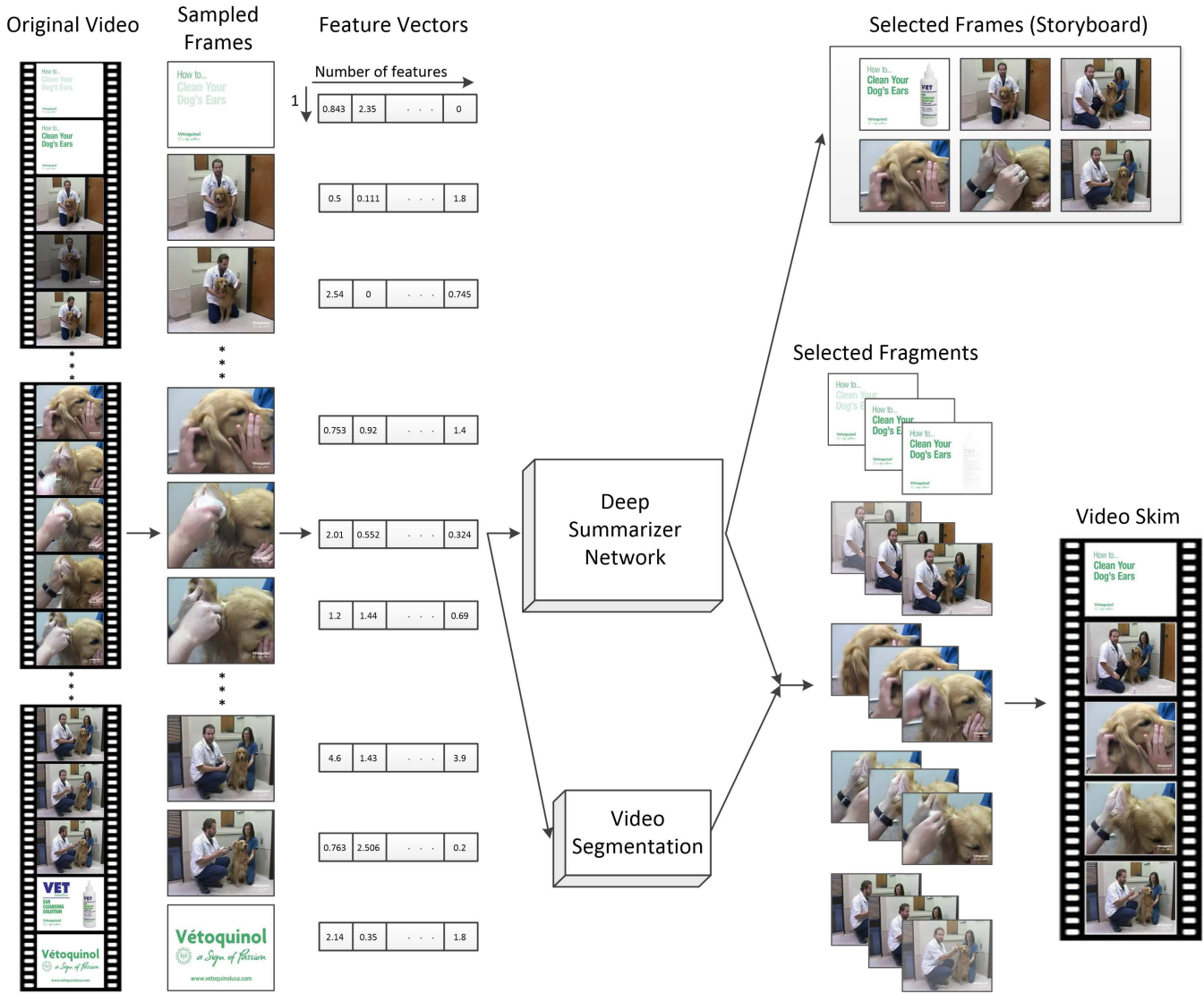}
\end{center}
   \caption{High-level representation of the analysis pipeline of the deep-learning-based video summarization methods for generating a video storyboard and a video skim.}
\label{fig:pipeline}
\end{figure*}

A high-level representation of the typical deep-learning-based video summarization pipeline is depicted in Fig.~\ref{fig:pipeline}. The first step of the analysis involves the representation of the visual content of the video with the help of feature vectors. Most commonly, such vectors are extracted at the frame-level, for all frames or for a subset of them selected via a frame-sampling strategy, e.g., processing $2$ frames per second. In this way, the extracted feature vectors store information at a very detailed level and capture the dynamics of the visual content that are of high significance when selecting the video parts that form the summary. Typically, in most deep-learning-based video summarization techniques the visual content of the video frames is represented by deep feature vectors extracted with the help of pre-trained neural networks. For this, a variety of Convolutional Neural Networks (CNNs) and Deep Convolutional Neural Networks (DCNNs) have been used in the bibliography, that include GoogleNet (Inception V1) \cite{7298594}, Inception V3 \cite{szegedy2016rethinking}, AlexNet \cite{krizhevsky2012imagenet}, variations of ResNet \cite{he2016deep} and variations of VGGnet \cite{simonyan2014very}. Nevertheless, the GoogleNet appears to be the most commonly used one thus far (used in \cite{10.1007/978-3-319-46478-7_47,Zhao:2017:HRN:3123266.3123328,10.1007/978-3-030-05716-9_6,10.1007/978-3-030-21074-8_4,8667390,JI2020200,10.1007/978-3-030-01258-8_22,Feng:2018:EVS:3240508.3240651,Wang:2019:SMN:3343031.3350992,8659061,Mahasseni2017UnsupervisedVS,Apostolidis:2019:SLA:3347449.3357482,DBLP:conf/wacv/FuTC19,jung2019discriminative,Yuan2019CycleSUMCA,10.1007/978-3-030-37731-1_40,He:2019:UVS:3343031.3351056,Rochan_2019_CVPR,Zhou2018DeepRL,gonuguntla2019enhanced,8924889,10.1007/978-3-030-01264-9_12,10.1145/3338533.3366583,Zhou2018VideoSB,8101557,9259058,10.1007/978-3-030-58595-2_11,DBLP:journals/mta/YalinizI21,LI2021107677,9037206}). The extracted features are then utilized by a deep summarizer network, which is trained by trying to minimize an objective function or a set of objective functions.

The output of the trained Deep Summarizer Network can be either a set of selected video frames (key-frames) that form a static video storyboard, or a set of selected video fragments (key-fragments) that are concatenated in chronological order and form a short video skim. With respect to the generated video storyboard, this should be similar with the sets of key-frames that would be selected by humans and must exhibit minimal visual redundancy. With regards to the produced video skim, this typically should be equal or less than a predefined length $L$. For experimentation and comparison purposes, this is most often set as $L = p \cdot T$ where $T$ is the video duration and $p$ is the ratio of the summary to video length; $p=0.15$ is a typical value, in which case the summary should not exceed $15\%$ of the original video's duration. As a side note, the production of a video skim (which is the ultimate goal of the majority of the proposed deep-learning-based summarization algorithms) requires the segmentation of the video into consecutive and non-overlapping fragments that exhibit visual and temporal coherence, thus offering a seamless presentation of a part of the story. Given this segmentation and the estimated frames' importance scores by the trained Deep Summarizer Network, video-segment-level importance scores are computed by averaging the importance scores of the frames that lie within each video segment. These segment-level scores are then used to select the key-fragments given the summary length $L$, and most methods (e.g., \cite{10.1007/978-3-319-46478-7_47,10.1007/978-3-030-05716-9_6,10.1007/978-3-030-21074-8_4,8667390,JI2020200,10.1007/978-3-030-01258-8_22,Feng:2018:EVS:3240508.3240651,lal2019online,chu2019spatiotemporal,8659061,Mahasseni2017UnsupervisedVS,Apostolidis:2019:SLA:3347449.3357482,Zhang:2019:DDT:3321408.3322622,10.1007/978-3-030-37731-1_40,He:2019:UVS:3343031.3351056,Rochan_2019_CVPR,Zhou2018DeepRL,gonuguntla2019enhanced,8924889,10.1145/3338533.3366583,8101557,9259058,DBLP:journals/mta/YalinizI21,LI2021107677}) tackle this step by solving the Knapsack problem.

With regards to the utilized type of data, the current bibliography on deep-learning-based video summarization can be divided between:
\begin{itemize}
    \item Unimodal approaches that utilize only the visual modality of the videos for feature extraction, and learn summarization in a (weakly-)supervised or unsupervised manner.
    \item Multimodal methods that exploit the available textual metadata and learn semantic/category-driven summarization in a supervised way by increasing the relevance between the semantics of the summary and the semantics of the associated metadata or video category.
\end{itemize}

Concerning the adopted training strategy, the existing deep-learning-based video summarization algorithms can be coarsely categorized in the following categories: 
\begin{itemize}
    \item Supervised approaches that rely on datasets with human-labeled ground-truth annotations (either in the form of video summaries, as in the case of the SumMe dataset \cite{10.1007/978-3-319-10584-0_33}, or in the form of frame-level importance scores, as in the case of the TVSum dataset \cite{7299154}), based on which they try to discover the underlying criterion for video frame/fragment selection and video summarization.
    \item Unsupervised approaches that overcome the need for ground-truth data (whose production requires time-demanding and laborious manual annotation procedures), based on learning mechanisms that require only an adequately large collection of original videos for their training.
    \item Weakly-supervised approaches that, similarly to unsupervised approaches, aim to alleviate the need for large sets of hand-labeled data. Less-expensive weak labels are utilized with the understanding that they are imperfect compared to a full set of human annotations, but can nonetheless be used to create strong predictive models.
\end{itemize}

\begin{figure*}[]
\begin{center}
   \includegraphics[width=0.73\textwidth]{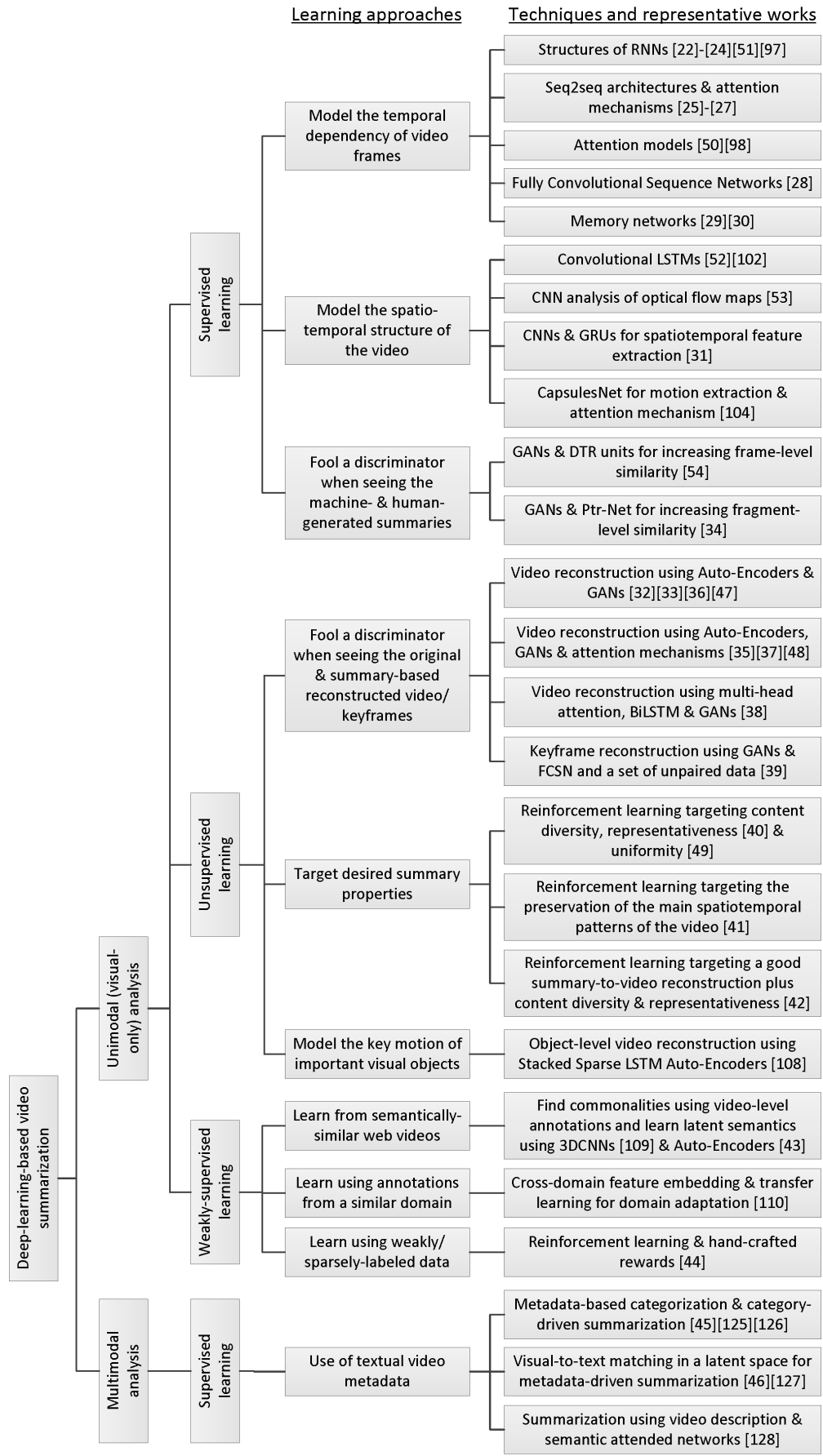}
\end{center}
   \caption{A taxonomy of the existing deep-learning-based video summarization algorithms.}
\label{fig:taxonomy}
\end{figure*}

Building on the above described categorizations, a more detailed taxonomy of the relevant bibliography is depicted in Fig. \ref{fig:taxonomy}. The penultimate layer of this arboreal illustration shows the different learning approaches that have been adopted. The leafs of each node of this layer show the utilized techniques for implementing each learning approach, and contain references to the most relevant works in the bibliography. This taxonomy will be the basis for presenting the relevant bibliography in the following section.

\section{Deep learning approaches}
\label{SoA}
This section gives a brief introduction to deep learning architectures (Section \ref{subsec:deep_learning_basics}), and then focuses on their application on the video summarization domain by providing a systematic review of the relevant bibliography. This review starts by presenting the different classes of supervised (Section \ref{subsec:supervised}), unsupervised (Section \ref{subsec:unsupervised}), and weakly-supervised (Section \ref{subsec:weakly-supervised}) video summarization methods, which rely solely on the analysis of the visual content. Following, it reports on multimodal approaches (Section \ref{subsec:multimodal}) that process also the available text-based metadata. Finally, it provides some general remarks (Section \ref{subsec:general_remarks}) that outline how the field has evolved, especially over the last five years.

\subsection{Deep Learning Basics}
\label{subsec:deep_learning_basics}
Deep learning is a branch of machine learning that was fueled by the explosive growth and availability of data and the remarkable advancements in hardware technologies. The term ``deep'' refers to the use of multiple layers in the network, that perform non-linear processing to learn multiple levels of data representations. Learning can be supervised, semi-supervised or unsupervised. Several deep learning architectures have been proposed thus far, that can be broadly classified in: Deep Belief Networks \cite{10.1162/neco.2006.18.7.1527}, Restricted/Deep Boltzmann Machines \cite{10.1145/1273496.1273596, pmlr-v5-salakhutdinov09a}, (Variational) Autoencoders \cite{10.5555/2987189.2987190, DBLP:journals/corr/KingmaW13}, (Deep) Convolutional Neural Networks \cite{10.5555/303568.303704}, Recursive Neural Networks \cite{548916}, Recurrent Neural Networks \cite{10.5555/65669.104451}, Generative Adversarial Networks \cite{10.5555/2969033.2969125}, Graph (Convolutional) Neural Networks \cite{4700287}, and Deep Probabilistic Neural Networks \cite{DBLP:conf/cvpr/Gast018}. For an overview of the different classes of deep learning architectures, the interested reader is referred to surveys such as \cite{LIU201711, 10.1145/3234150, DONG2021100379}. Over the last decade, deep learning architectures have been used in several applications, including natural language processing (e.g., \cite{schwenk-2012-continuous,10.1109/TASLP.2015.2509257}), speech recognition (e.g., \cite{7230356,10.1016/j.csl.2016.06.007}), medical image/video analysis (e.g., \cite{7399414,10.1007/978-3-030-59716-0_46}) and computer vision (e.g., \cite{7780459,DBLP:journals/corr/SimonyanZ14a,8578223,10.1145/2647868.2654914}), leading to state-of-the-art results and performing in many cases comparably to a human expert. For additional information about applications of deep learning we refer the reader to the recent surveys \cite{GU2018354,BOUWMANS20198,DIXIT2021100317,10.3389/fnins.2020.00779,9068523,9039685,rob.21918}.

Nevertheless, the empirical success of deep learning architectures is associated with numerous challenges for the theoreticians, which are critical to the training of deep networks. Such challenges relate to: i) the design of architectures that are able to learn from sparse, missing or noisy training data, ii) the use of optimization algorithms to adjust the network parameters, iii) the implementation of compact deep network architectures that can be integrated into mobile devices with restricted memory, iv) the analysis of the stability of deep networks, and v) the explanation of the underlying mechanisms that are activated at inference time in a way that is easily understandable by humans (the relevant research domain is also known as Explainable AI). Such challenges and some suggested approaches to addressing them are highlighted in several works, e.g., \cite{10.5555/3327546.3327690,8438540,NIPS2013_3871bd64,alain2018understanding,7410686,Yun2018ACV,10.5555/3327757.3327888}.

\subsection{Supervised Video Summarization}
\label{subsec:supervised}

\begin{figure}[t]
\begin{center}
   \includegraphics[width=0.99\columnwidth]{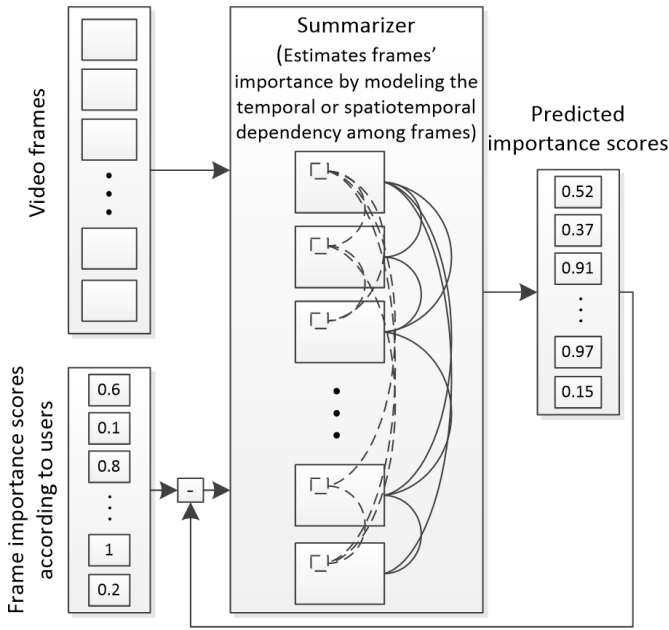}
\end{center}
   \caption{High-level representation of the analysis pipeline of supervised algorithms that perform summarization by learning the frames' importance after modeling their temporal or spatiotemporal dependency. For the latter class of methods (i.e., modeling the spatiotemporal dependency among frames), object bounding boxes and object relations in time shown with dashed rectangles and lines, are used to illustrate the extension that models both the temporal and spatial dependency among frames.}
\label{fig:sup2_4}
\end{figure}

\textbf{1. Learn frame importance by modeling the temporal dependency among frames.}
Early deep-learning-based approaches cast summarization as a structured prediction problem and try to make estimates about the frames' importance by modeling their temporal dependency. As illustrated in Fig. \ref{fig:sup2_4}, during the training phase the Summarizer gets as input the sequence of the video frames and the available ground-truth data that indicate the importance of each frame according to the users' preferences. These data are then used to model the dependencies among the video frames in time (illustrated with solid arched lines) and estimate the frames' importance. The predicted importance scores are compared with the ground-truth data and the outcome of this comparison guides the training of the Summarizer. The first approach to this direction, proposed by Zhang et al. (2016) \cite{10.1007/978-3-319-46478-7_47}, uses Long Short-Term Memory (LSTM) units \cite{hochreiter1997} to model variable-range temporal dependency among video frames. Frames' importance is estimated using a multi-layer perceptron (MLP), and the diversity of the visual content of the generated summary is increased based on the Determinantal Point Process (DPP) \cite{10.5555/2481023}. One year later, Zhao et al. (2017) \cite{Zhao:2017:HRN:3123266.3123328} described a two-layer LSTM architecture. The first layer extracts and encodes data about the video structure. The second layer uses this information to estimate fragment-level importance and select the key-fragments of the video. In their subsequent work, Zhao et al. (2018) \cite{ZhaoHSARNNHS} integrated a component that is trained to identify the shot-level temporal structure of the video. This knowledge is then utilized for estimating importance at the shot-level and producing a key-shot-based video summary. In their last work, Zhao et al. (2020) \cite{9037206} extended the method of \cite{Zhao:2017:HRN:3123266.3123328} by introducing a tensor-train embedding layer to avoid large feature-to-hidden mapping matrices. This layer is combined with a hierarchical structure of RNNs, that operates similarly to the one in \cite{Zhao:2017:HRN:3123266.3123328} and captures the temporal dependency of frames that lie within manually-defined videos subshots (first layer) and over the different subshots of the video (second layer). The output of these layers is used for determining the probability of each subshot to be selected as a part of the video summary. Lebron Casas et al. (2019) \cite{10.1007/978-3-030-05716-9_6} built on \cite{10.1007/978-3-319-46478-7_47} and introduced an attention mechanism to model the temporal evolution of the users' interest. Following, this information is used to estimate frames' importance and select the video key-frames to build a video storyboard. In the same direction, a few methods utilized sequence-to-sequence (a.k.a. seq2seq) architectures in combination with attention mechanisms. Ji et al. (2020) \cite{8667390} formulated video summarization as a seq2seq learning problem and proposed an LSTM-based encoder-decoder network with an intermediate attention layer. Ji et al. (2020) \cite{JI2020200} introduced an extension of their summarization model from \cite{8667390}, which integrates a semantic preserving embedding network that evaluates the output of the decoder with respect to the preservation of the video's semantics using a tailored semantic preserving loss, and replacing the previously used Mean Square Error (MSE) loss by the Huber loss to enhance its robustness to outliers. Using the attention mechanism as the core part of the analysis and aiming to avoid the use of computationally-demanding LSTMs, Fajtl et al. (2019) \cite{10.1007/978-3-030-21074-8_4} presented a network for video summarization, that is composed of a soft, self-attention mechanism and a two-layer fully connected network for regression of the frames' importance scores. Liu et al. (2019) \cite{liu2019learning} described a hierarchical approach which combines a generator-discriminator architecture (similar to the one in \cite{Mahasseni2017UnsupervisedVS}) as an internal mechanism to estimate the representativeness of each shot and define a set of candidate key-frames. Then, it employs a multi-head attention model to further assess candidates' importance and select the key-frames that form the summary. Li et al. (2021) \cite{LI2021107677} proposed a global diverse attention mechanism by making an adaptation of the self-attention mechanism of the Transformer Network \cite{10.5555/3295222.3295349}. This mechanism is based on a pairwise similarity matrix that contains diverse attention weights for the video frames and encodes temporal relations between every two frames in a wide range of stride. The estimated diverse attention weights are then transformed to importance scores through a regression mechanism, and these scores are compared with ground-truth annotations to learn video summarization in a supervised manner. Following another approach to model the dependency of video frames, Rochan et al. (2018) \cite{10.1007/978-3-030-01258-8_22} tackled video summarization as a semantic segmentation task where the input video is seen as a 1D image (of size equal to the number of video frames) with K channels that correspond to the K dimensions of the frames' representation vectors (either containing raw pixel values or being precomputed feature vectors). Then, they used popular semantic segmentation models, such as Fully Convolutional Networks (FCN) \cite{7298965} and an adaptation of DeepLab \cite{7913730}, and built a network (called Fully Convolutional Sequence Network) for video summarization. The latter consists of a stack of convolutions with increasing effective context size as we go deeper in the network, that enable the network to effectively model long-range dependency among frames and learn frames' importance. Finally, to address issues related to the limited capacity of LSTMs, some techniques use additional memory. Feng et al. (2018) \cite{Feng:2018:EVS:3240508.3240651} described a deep learning architecture that stores information about the entire video in an external memory and predicts each shot's importance by learning an embedding space that enables matching of each shot with the entire memory information. In a more recent method, Wang et al. (2019) \cite{Wang:2019:SMN:3343031.3350992} stacked multiple LSTM and memory layers hierarchically to derive long-term temporal context, and used this information to estimate the frames' importance.

\textbf{2. Learn frame importance by modeling the spatiotemporal structure of the video.}
Aiming to learn how to make better estimates about the importance of video frames/fragments, some techniques pay attention to both the spatial and temporal structure of the video. Again, the Summarizer gets as input the sequence of the video frames and the available ground-truth data that indicate the importance of each frame according to the users' preferences. But, extending the analysis pipeline of the previously described group of methods, it then also models the spatiotemporal dependencies among frames (shown with dashed rectangles and lines in Fig. \ref{fig:sup2_4}). Once again, the predicted importance scores are compared with the ground-truth data and the outcome of this comparison guides the training of the Summarizer. From this perspective, Lal et al. (2019) \cite{lal2019online} presented an encoder-decoder architecture with convolutional LSTMs, that models the spatiotemporal relationship among parts of the video. In addition to the estimates about the frames' importance, the algorithm enhances the visual diversity of the summary via next frame prediction and shot detection mechanisms, based on the intuition that the first frames of a shot generally have high likelihood of being part of the summary. Yuan et al. (2019) \cite{yuan2019spatiotemporal} extracted deep and shallow features from the video content using a trainable 3D-CNN and built a new representation through a fusion strategy. Then, they used this representation in combination with convolutional LSTMs to model the spatial and temporal structure of the video. Finally, summarization is learned with the help of a new loss function (called Sobolev loss) that aims to define a series of frame-level importance scores that is close to the series of ground-truth scores by minimizing the distance of the derivatives of these sequential data, and to exploit the temporal structure of the video. Chu et al. (2019) \cite{chu2019spatiotemporal} extracted spatial and temporal information by processing the raw frames and their optical flow maps with CNNs, and learned how to estimate frames' importance based on human annotations and a label distribution learning process. Elfeki et al. (2019) \cite{8659061} combined CNNs and Gated Recurrent Units \cite{cho-etal-2014-learning} (a type of RNN) to form spatiotemporal feature vectors, that are then used to estimate the level of activity and importance of each frame. Huang et al. (2020) \cite{8601376} trained a neural network for spatiotemporal data extraction and used the extracted information to create an inter-frames motion curve. The latter is utilized as input to a transition effects detection method that segments the video into shots. Finally, a self-attention model exploits the human-generated ground-truth data to learn how to estimate the intra-shot importance and select the key-frames/fragments of the video to form a static/dynamic video summary.

\begin{figure}[t]
\begin{center}
   \includegraphics[width=0.93\columnwidth]{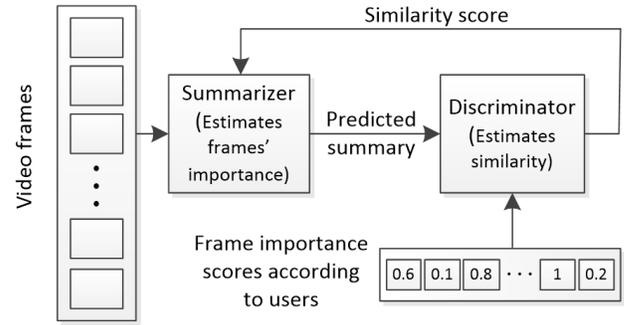}
\end{center}
   \caption{High-level representation of the analysis pipeline of supervised algorithms that learn summarization with the help of ground-truth data and adversarial learning.}
\label{fig:sup3}
\end{figure}

\textbf{3. Learn summarization by fooling a discriminator when trying to discriminate a machine-generated from a human-generated summary.}
Following a completely different approach to minimizing the distance between the machine-generated and the ground-truth summaries, a couple of methods use Generative Adversarial Networks (GANs). As presented in Fig. \ref{fig:sup3}, the Summarizer (which acts as the Generator of the GAN) gets as input the sequence of the video frames and generates a summary by computing frame-level importance scores. The generated summary (i.e., the predicted frame-level importance scores) along with an optimal video summary for this video (i.e., frame-level importance scores according to the users' preferences) are given as input to a trainable Discriminator which outputs a score that quantifies their similarity. The training of the entire summarization architecture is performed in an adversarial manner. The Summarizer tries to fool the Discriminator to not distinguish the predicted from the user-generated summary, and the Discriminator aims to learn how to make this distinction. When the Discriminator's confidence is very low (i.e., the classification error is approximately equal for both the machine- and the user-generated summary), then the Summarizer is able to generate a summary that is very close to the users' expectations. In this context, Zhang et al. (2019) \cite{Zhang:2019:DDT:3321408.3322622} proposed a method that combines LSTMs and Dilated Temporal Relational (DTR) units to estimate temporal dependencies among frames at different temporal windows, and learns summarization by trying to fool a trainable discriminator when distinguishing the machine-based summary from the ground-truth and a randomly-created one. In another work from the same year, Fu et al. (2019) \cite{DBLP:conf/wacv/FuTC19} suggested an adversarial learning approach for (semi-)supervised video summarization. The Generator/Summarizer is an attention-based Pointer Network \cite{NIPS2015_5866} that defines the start and end point of each video fragment that is used to form the summary. The Discriminator is a 3D-CNN classifier that judges whether a fragment is from a ground-truth or a machine-generated summary. Instead of using the typical adversarial loss, in this algorithm the output of the Discriminator is used as a reward to train the Generator/Summarizer based on reinforcement learning. Thus far, the use of GANs for supervised video summarization is limited. Nevertheless, this machine learning framework has been widely used for unsupervised video summarization, as discussed in the following section. 

\subsection{Unsupervised Video Summarization}
\label{subsec:unsupervised}

\begin{figure}[t]
\begin{center}
   \includegraphics[width=\columnwidth]{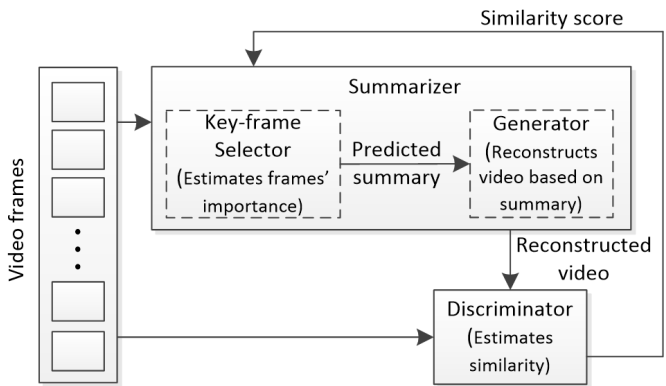}
\end{center}
   \caption{High-level representation of the analysis pipeline of unsupervised algorithms that learn summarization by increasing the similarity between the summary and the video.}
\label{fig:unsup1}
\end{figure}

\textbf{1. Learn summarization by fooling a discriminator when trying to discriminate the original video (or set of key-frames) from a summary-based reconstruction of it.}
Given the lack of any guidance (in the form of ground-truth data) for learning video summarization, most existing unsupervised approaches rely on the rule that a representative summary ought to assist the viewer to infer the original video content. In this context, these techniques utilize GANs to learn how to create a video summary that allows a good reconstruction of the original video. The main concept of this training approach is depicted in Fig. \ref{fig:unsup1}. The Summarizer is usually composed of a Key-frame Selector that estimates the frames' importance and generates a summary, and a Generator that reconstructs the video based on the generated summary. It gets as input the sequence of the video frames and, through the aforementioned internal processing steps, reconstructs the original video based on the generated summary (which is represented by the predicted frame-level importance scores). The reconstructed video along with the original one are given as input to a trainable Discriminator which outputs a score that quantifies their similarity. Similarly to the supervised GAN-based methods, the training of the entire summarization architecture is performed in an adversarial manner. However, in this case the Summarizer tries to fool the Discriminator when distinguishing the summary-based reconstructed video from the original one, while the Discriminator aims to learn how to make the distinction. When this discrimination is not possible (i.e., the classification error is approximately equal for both the reconstructed and the original video), the Summarizer is considered to be able to build a video summary that is highly-representative of the overall video content. To this direction, the work of Mahasseni et al. (2017) \cite{Mahasseni2017UnsupervisedVS} is the first that combines an LSTM-based key-frame selector with a Variational Auto-Encoder (VAE) and a trainable Discriminator, and learns video summarization through an adversarial learning process that aims to minimize the distance between the original video and the summary-based reconstructed version of it. Apostolidis et al. (2019) \cite{Apostolidis:2019:SLA:3347449.3357482} built on the network architecture of \cite{Mahasseni2017UnsupervisedVS} and suggested a stepwise, label-based approach for training the adversarial part of the network, that leads to improved summarization performance. Yuan et al. (2019) \cite{Yuan2019CycleSUMCA} proposed an approach that aims to maximize the mutual information between the summary and the video using a trainable couple of discriminators and a cycle-consistent adversarial learning objective. The frame selector (bi-directional LSTM) builds a video summary by modeling the temporal dependency among frames. This summary is then forwarded to the evaluator which is composed of two GANs; the forward GAN is used to learn how to reconstruct the original video from the video summary, and the backward GAN tries to learn how to perform the backward reconstruction from the original to the summary video. The consistency between the output of such cycle learning is used as a measure that quantifies information preservation between the original video and the generated summary. Using this measure, the evaluator guides the frame selector to identify the most informative frames and form the video summary. In one of their subsequent works, Apostolidis et al. (2020) \cite{9259058} embedded an Actor-Critic model into a GAN and formulated the selection of important video fragments (that will be used to form the summary) as a sequence generation task. The Actor and the Critic take part in a game that incrementally leads to the selection of the video key-fragments, and their choices at each step of the game result in a set of rewards from the Discriminator. The designed training workflow allows the Actor and Critic to discover a space of actions and automatically learn a value function (Critic) and a policy for key-fragment selection (Actor). On the same direction, some approaches extended the core component of the aforementioned works (i.e., the VAE-GAN architecture) by introducing tailored attention mechanisms. Jung et al. (2019) \cite{jung2019discriminative} proposed a VAE-GAN architecture that is extended by a chunk and stride network (CSNet) and a tailored difference attention mechanism for assessing the frames' dependency at different temporal granularities when selecting the video key-frames. In their next work, Jung et al. (2020) \cite{10.1007/978-3-030-58595-2_11} introduced another approach for estimating frames' importance, that uses a self-attention mechanism (similar to the one integrated in the Transformer Network \cite{10.5555/3295222.3295349}) in combination with an algorithm for modeling the relative position between frames. The frame sequence is decomposed into equally-sized, non-overlapping groups of consecutive and neighboring frames (selected using a constant sampling step), to capture both the local and the global interdependencies between video frames. The proposed approach was considered as a strategy from estimating frames' importance and its effectiveness was evaluated after being integrated into the network architecture of \cite{jung2019discriminative}. Apostolidis et al. (2020) \cite{10.1007/978-3-030-37731-1_40} introduced a variation of their previous work \cite{Apostolidis:2019:SLA:3347449.3357482}, that replaces the Variational Auto-Encoder with a deterministic Attention Auto-Encoder for learning an attention-driven reconstruction of the original video, which subsequently improves the key-fragment selection process. He et al. (2019) \cite{He:2019:UVS:3343031.3351056} presented a self-attention-based conditional GAN. The Generator produces weighted frame features and predicts frame-level importance scores, while the Discriminator tries to distinguish between the weighted and the raw frame features. A conditional feature selector is used to guide the GAN model to focus on more important temporal regions of the whole video frames, while long-range temporal dependencies along the whole video sequence are modeled by a multi-head self-attention mechanism. Finally, building on a Generator-Discriminator mechanism, Rochan et al. (2019) \cite{Rochan_2019_CVPR} proposed an approach that learns video summarization from unpaired data based on an adversarial process that relies on GANs and a Fully-Convolutional Sequence Network (FCSN) encoder-decoder. The model of \cite{Rochan_2019_CVPR} aims to learn a mapping function of a raw video to a human-like summary, such that the distribution of the generated summary is similar to the distribution of human-created summaries, while content diversity is forced by applying a relevant constraint on the learned mapping function.

\begin{figure}[t]
\begin{center}
   \includegraphics[width=0.98\columnwidth]{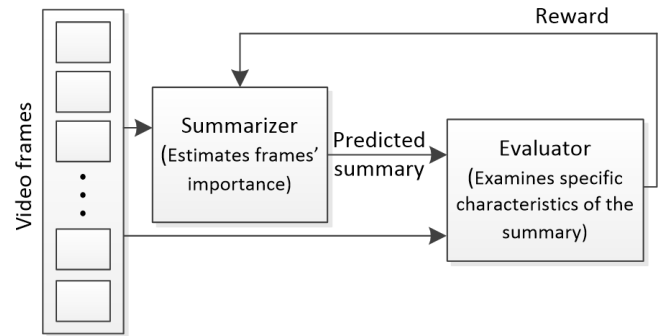}
\end{center}
   \caption{High-level representation of the analysis pipeline of supervised algorithms that learn summarization based on hand-crafted rewards and reinforcement learning.}
\label{fig:unsup2}
\end{figure}

\textbf{2. Learn summarization by targeting specific desired properties for the summary.}
Aiming to deal with the unstable training \cite{Zhou2018DeepRL} and the restricted evaluation criteria of GAN-based methods (that mainly focus on the summary's ability to lead to a good reconstruction of the original video), some unsupervised approaches perform summarization by targeting specific properties of an optimal video summary. To this direction, they utilize the principles of reinforcement learning in combination with hand-crafted reward functions that quantify the existence of desired characteristics in the generated summary. As presented in Fig. \ref{fig:unsup2}, the Summarizer gets as input the sequence of the video frames and creates a summary by predicting frame-level importance scores. The created (predicted) summary is then forwarded to an Evaluator which is responsible to quantify the existence of specific desired characteristics with the help of hand-crafted reward functions. The computed score(s) are then combined to form an overall reward value, that is finally used to guide the training of the Summarizer. The first work to this direction, proposed by Zhou et al. (2018) \cite{Zhou2018DeepRL}, formulates video summarization as a sequential decision-making process and trains a Summarizer to produce diverse and representative video summaries using a diversity-representativeness reward. The diversity reward measures the dissimilarity among the selected key-frames and the representativeness reward computes the distance (expressing the visual resemblance) of the selected key-frames from the remaining frames of the video. Building on this method, Yaliniz et al. (2021) \cite{DBLP:journals/mta/YalinizI21} presented another reinforcement-learning-based approach that considers also the uniformity of the generated summary. The temporal dependency among frames is modeled using Independently Recurrent Neural Networks (IndRNNs) \cite{Li2018IndependentlyRN} activated by a Leaky ReLU (Leaky Rectified Linear Unit) function; in this way, Yaliniz et al. try to overcome identified issues of LSTMs with regards to decaying, vanishing and exploding gradients, and better learn long term dependencies over the sequence of video frames. Moreover, besides using rewards associated with the representativeness and diversity of the video summary, to avoid redundant jumps between the selected video fragments that form the summary Yaliniz et al. added a uniformity reward that aims to enhance the coherence of the generated summary. Gonuguntla et al. (2019) \cite{gonuguntla2019enhanced} built a method that utilizes Temporal Segment Networks (proposed in \cite{10.1007/978-3-319-46484-8_2} for action recognition in videos) to extract spatial and temporal information about the video frames, and trains the Summarizer through a reward function that assesses the preservation of the video's main spatiotemporal patterns in the produced summary. Finally, Zhao et al. (2020) \cite{8924889} presented a mechanism for both video summarization and reconstruction. Video reconstruction aims to estimate the extent to which the summary allows the viewer to infer the original video (similarly to some of the above-presented GAN-based methods), and video summarization is learned based on the reconstructor's feedback and the output of trained models that assess the representativeness and diversity of the visual content of the generated summary. 

\textbf{3. Build object-oriented summaries by modeling the key-motion of important visual objects.}
Building on a different basis, Zhang et al. (2018) \cite{ZHANG2018} developed a method that focuses on the preservation in the summary of the underlying fine-grained semantic and motion information of the video. For this, it performs a preprocessing step that aims to find important objects and their key-motions. Based on this step, it represents the whole video by creating super-segmented object motion clips. Each one of these clips is then given to the Summarizer, which uses an online motion auto-encoder model (Stacked Sparse LSTM Auto-Encoder) to memorize past states of object motions by continuously updating a tailored recurrent auto-encoder network. The latter is responsible for reconstructing object-level motion clips, and the reconstruction loss between the input and the output of this component is used to guide the training of the Summarizer. Based on this training process the Summarizer is able to generate summaries that show the representative objects in the video and the key-motions of each of these objects.

\subsection{Weakly-supervised Video Summarization}
\label{subsec:weakly-supervised}

Weakly-supervised video summarization methods try to mitigate the need for extensive human-generated ground-truth data, similarly to unsupervised learning methods. Instead of not using any ground-truth data, they use less-expensive weak labels (such as video-level metadata for video categorization and category-driven summarization, or ground-truth annotations for a small subset of video frames for learning summarization through sparse reinforcement learning and tailored reward functions) under the main hypothesis that these labels are imperfect compared to a full set of human annotations, but can nonetheless allow the training of effective summarization models. We avoid the illustration of a typical analysis pipeline for this class of methods, as there is a limited overlap in the way that these methods conceptualize the learning of the summarization task. The first approach that adopts an intermediate way between fully-supervised and fully-unsupervised learning for video summarization was described by Panda et al. (2018) \cite{8237657}. This approach uses video-level metadata (e.g., the video title ``A man is cooking'') to define a categorization of videos. Then, it leverages multiple videos of a category and extracts 3D-CNN features to automatically learn a parametric model for categorizing new (unseen during training) videos. Finally, it adopts the learned model to select the video segments that maximize the relevance between the summary and the video category. Panda et al. investigated different ways to tackle issues related to the limited size of available datasets, that include cross-dataset training, the use of web-crawled videos, as well as data augmentation methods for obtaining sufficient training data. Building on the concept of learning summarization from semantically-similar videos, Cai et al. (2018) \cite{10.1007/978-3-030-01264-9_12} suggested a weakly-supervised setting of learning summarization models from a large number of web videos. Their architecture combines a Variational Auto-Encoder that learns the latent semantics from web videos, and a sequence encoder-decoder with attention mechanism that performs summarization. The decoding part of the VAE aims to reconstruct the input videos using samples from the learned latent semantics; while, the most important video fragments are identified through the soft attention mechanism of the encoder-decoder network, where the attention vectors of raw videos are obtained by integrating the learned latent semantics from the collected web videos. The overall architecture is trained by a weakly-supervised semantic matching loss to learn the topic-associated summaries. Ho et al. (2018) \cite{10.1007/978-3-030-01267-0_5} proposed a deep learning framework for summarizing first-person videos; however, we report on this method here, as it is also evaluated on a dataset used to assess generic video summarization methods. Given the observation in \cite{10.1007/978-3-030-01267-0_5} that the collection of a sufficiently large amount of fully-annotated first-person video data with ground-truth annotations is a difficult task, Ho et al. built an algorithm that exploits the principles of transfer learning and uses annotated third-person videos (which, as argued in \cite{10.1007/978-3-030-01267-0_5}, can be found more easily) to learn how to summarize first-person videos. The algorithm performs cross-domain feature embedding and transfer learning for domain adaptation (across third- and first-person videos) in a semi-supervised manner. In particular, training is performed based on a set of third-person videos with fully annotated highlight scores and a set of first-person videos where only a small portion of them comes with ground-truth scores. Finally, Chen et al. (2019) \cite{10.1145/3338533.3366583} utilized the principles of reinforcement learning to build and train a summarization method based on a limited set of human annotations and a set of hand-crafted rewards. The latter relate to the similarity between the machine- and the human-selected fragments, as well as to specific characteristics of the created summary (e.g., its representativeness). More specifically, this method applies a hierarchical key-fragment selection process that is divided into sub-tasks. Each task is learned through sparse reinforcement learning (thus avoiding the need for exhaustive annotations about the entire set of frames, and using annotations only for a subset of frames) and the final summary is formed based on rewards about its diversity and representativeness.

\subsection{Multimodal Approaches}
\label{subsec:multimodal}

A number of works investigated the potential of exploiting additional modalities (besides the visual stream) for learning summarization, such as the audio stream, the video captions or ASR transcripts, any available textual metadata (video title and/or abstract) or other contextual data (e.g., viewers' comments). Several of these multimodal approaches were proposed before the so-called ``deep learning era'', targeting either generic or domain/task-specific video summarization. Some indicative and recent examples can be found in \cite{10.1007/978-3-642-11301-7_40,5993544,6682798,10.1145/2072298.2072068,7023833,6527322,7351630,8099601}. Addressing the video summarization task from a slightly different perspective, other multimodal algorithms generate a text-based summary of the video \cite{li-etal-2017-multi,libovicky2018multimodal,palaskar-etal-2019-multimodal}, and extend this output by extracting one representative key-frame \cite{zhu-etal-2018-msmo}. A multimodal deep-learning-based algorithm for summarizing videos of soccer games was presented in \cite{10.1145/3347318.3355524}, while another multimodal approach for key-frame extraction from first-person videos that exploits sensor data was described in \cite{8297032}. Nevertheless, all these works are outside the scope of this survey, which focuses on deep-learning-based methods for generic video summarization; i.e., methods that learn summarization with the help of deep neural network architectures and/or the visual content is represented by deep features.

\begin{figure}[t]
\begin{center}
   \includegraphics[width=0.95\columnwidth]{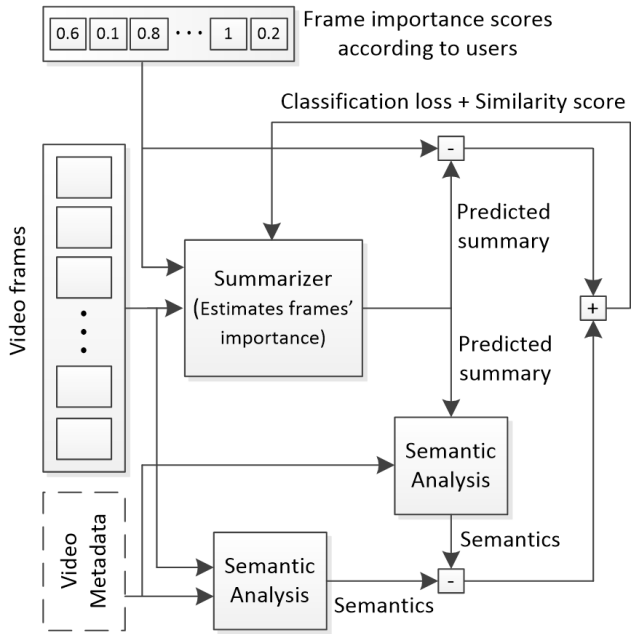}
\end{center}
   \caption{High-level representation of the analysis pipeline of supervised algorithms that perform semantic/category-driven summarization.}
\label{fig:sup1}
\end{figure}

The majority of multimodal approaches that are within the focus of this study tackle video summarization by utilizing also the textual video metadata, such as the video title and description. As depicted in Fig. \ref{fig:sup1}, during training the Summarizer gets as input: i) the sequence of the video frames that needs to be summarized, ii) the relevant ground-truth data (i.e., frame-level importance scores according to the users), and iii) the associated video metadata. Following, it estimates the frames' importance and generates (predicts) a summary. Then, the generated summary is compared with the ground-truth summary and the video metadata. The former comparison produces a similarity score. The latter comparison involves the semantic analysis of the summary and the video metadata. Given the output of this analysis, most algorithms try to minimize the distance of the generated representations in a learned latent space, or the classification loss when the aim is to define a summary that maintains the core characteristics of the video category. The combined outcome of this assessment is finally used to guide the training of the Summarizer. In this context, Song et al. (2016) \cite{7574720} and Zhou et al. (2018) \cite{Zhou2018VideoSB} proposed methods that learn category-driven summarization by rewarding the preservation of the core parts found in video summaries from the same category (e.g., the main parts of a wedding ceremony when summarizing a wedding video). On the same direction, Lei et al. (2019) \cite{8421619} presented a method which uses action classifiers that have been trained with video-level labels, to perform action-driven video fragmentation and labeling; then, this method extracts a fixed number of key-frames and applies reinforcement learning to select the ones with the highest categorization accuracy, thus performing category-driven summarization. Building on the idea of exploiting contextual data, Otani et al. (2017) \cite{10.1007/978-3-319-54193-8_23} and Yuan et al. (2019) \cite{8101557}, suggested methods that define a video summary by maximizing its relevance with the available video metadata, after projecting both the visual and the textual information in a common latent space. Finally, approaching the summarization task from a different perspective, Wei et al. (2018) \cite{Wei2018VideoSV} introduced an approach that applies a visual-to-text mapping and a semantic-based video fragment selection process according to the relevance between the automatically-generated and the original video description, with the help of semantic attended networks. However, most of these methods examine only the visual cues without considering the sequential structure of the video and the temporal dependency among frames.

\subsection{General remarks on deep learning approaches}
\label{subsec:general_remarks}

Based on the review of the relevant bibliography, we saw that early deep-learning-based approaches for video summarization utilize combinations of CNNs and RNNs. The former are used as pre-trained components (e.g., CNNs trained on ImageNet for visual concept detection) to represent the visual content, and the latter (mostly LSTMs) are used to model the temporal dependency among video frames. The majority of these approaches are supervised and try to learn how to make estimates about the importance of video frames/fragments based on human-generated ground-truth annotations. Architectures of RNNs can be used in the typical or in a hierarchical form \cite{10.1007/978-3-319-46478-7_47,Zhao:2017:HRN:3123266.3123328,ZhaoHSARNNHS,9037206} to also model the temporal structure and utilize this knowledge when selecting the video fragments of the summary. In some cases such architectures are combined with attention mechanisms to model the evolution of the users' interest \cite{10.1007/978-3-030-05716-9_6,8667390,JI2020200}, or extended by memory networks to increase the memorization capacity of the architecture and capture long-range temporal dependencies among parts of the video \cite{Feng:2018:EVS:3240508.3240651,Wang:2019:SMN:3343031.3350992}. Alternatively, some works \cite{10.1007/978-3-030-21074-8_4,LI2021107677} avoid the use of computationally-demanding RNNs, and instead they model the frames' dependencies with the help of learnable similarity-based attention mechanisms. Going one step further, a group of techniques try to learn importance by paying attention to both the spatial and temporal structure of the video, using convolutional LSTMs \cite{lal2019online,yuan2019spatiotemporal}, optical flow maps \cite{chu2019spatiotemporal}, combinations of CNNs and RNNs \cite{8659061}, or motion extraction mechanisms \cite{8601376}. Following a different approach, a couple of supervised methods learn how to generate video summaries that are aligned with the human preferences with the help of GANs \cite{Zhang:2019:DDT:3321408.3322622,DBLP:conf/wacv/FuTC19}. Finally, multimodal algorithms extract the high-level semantics of the visual content using pre-trained CNNs/DCNNs and learn summarization in a supervised manner by maximizing the semantic similarity among the visual summary and the contextual video metadata or the video category. However, the latter methods focus mostly on the visual content and disregard the sequential nature of the video, which is essential when summarizing the presented story.

To train a video summarization network in a fully unsupervised manner, the use of GANs seems to be the central direction thus far \cite{Mahasseni2017UnsupervisedVS,Apostolidis:2019:SLA:3347449.3357482,jung2019discriminative,Yuan2019CycleSUMCA,10.1007/978-3-030-37731-1_40,He:2019:UVS:3343031.3351056,Rochan_2019_CVPR,9259058}. The main intuition behind the use of GANs is that the produced summary should allow the viewer to infer the original video, and thus the unsupervised GAN-based methods are trying to build a summary that enables a good reconstruction of the original video. In most cases the generative part is composed of an LSTM that estimates the frames' importance according to their temporal dependency, thus indicating the most appropriate video parts for the summary. Then, the reconstruction of the video based on the predicted summary is performed with the help of Auto-Encoders \cite{Mahasseni2017UnsupervisedVS,Apostolidis:2019:SLA:3347449.3357482,Yuan2019CycleSUMCA} that in some cases are combined with tailored attention mechanisms \cite{jung2019discriminative,10.1007/978-3-030-37731-1_40}. Alternatively, the selection of the most important frames or fragments can be assisted by the use of Actor-Critic models \cite{9259058} or Transformer-like self-attention mechanisms \cite{10.1007/978-3-030-58595-2_11}. Another, but less popular, approach for unsupervised video summarization is the use of reinforcement learning in combination with hand-crafted rewards about specific properties of the generated summary. These rewards usually aim to increase the representativeness, diversity \cite{Zhou2018DeepRL} and uniformity \cite{DBLP:journals/mta/YalinizI21} of the summary, retain the statiotemporal patterns of the video \cite{gonuguntla2019enhanced}, or secure a good summary-based reconstruction of the video \cite{8924889}. Finally, one unsupervised method learns how to build object-oriented summaries by modeling the key-motion of important visual objects using a stacked sparse LSTM Auto-Encoder \cite{ZHANG2018}.

Last but not least, a few weakly-supervised methods have also been proposed. These methods learn video summarization by exploiting the semantics of the video metadata \cite{8237657} or the summary structure of semantically-similar web videos \cite{10.1007/978-3-030-01264-9_12}, exploiting annotations from a similar domain and transferring the gained knowledge via cross-domain feature embedding and transfer learning techniques \cite{10.1007/978-3-030-01267-0_5}, or using weakly/sparsely-labeled data under a reinforcement learning framework \cite{10.1145/3338533.3366583}.

With respect to the potential of deep-learning-based video summarization algorithms, we argue that, despite the fact that currently the major research direction is towards the development of supervised algorithms, the exploration of the learning capability of fully-unsupervised and semi/weakly-supervised methods is highly recommended. The reasoning behind this claim is based on the fact that: i) the generation of ground-truth training data (summaries) can be an expensive and laborious process; ii) video summarization is a subjective task, thus multiple different summaries can be proposed for a video from different human annotators; and iii) these ground-truth summaries can vary a lot, thus making it hard to train a method with the typical supervised training approaches. On the other hand, unsupervised video summarization algorithms overcome the need for ground-truth data as they can be trained using only an adequately large collection of original, full-length videos. Moreover, unsupervised and semi/weakly-supervised learning allows to easily train different models of the same deep network architecture using different types of video content (TV shows, news) and user-specified rules about the content of the summary, thus facilitating the domain-adaptation of video summarization. Given the above, we believe that unsupervised and semi/weakly-supervised video summarization have great advantages, and thus their potential should be further investigated.

\section{Evaluating video summarization}
\label{evaluation}
\subsection{Datasets}
Four datasets prevail in the video summarization bibliography: SumMe \cite{10.1007/978-3-319-10584-0_33}, TVSum \cite{7299154}, OVP \cite{10.1016/j.patrec.2010.08.004} and Youtube \cite{10.1016/j.patrec.2010.08.004}. SumMe consists of 25 videos of 1 to 6 minutes duration, with diverse video contents, captured from both first-person and third-person view. Each video has been annotated by 15 to 18 users in the form of key-fragments, and thus is associated to multiple fragment-level user summaries that have a length between 5\% and 15\% of the initial video duration. TVSum consists of 50 videos of 1 to 11 minutes duration, containing video content from 10 categories
of the TRECVid MED dataset. The TVSum videos have been
annotated by 20 users in the form of shot- and frame-level importance scores (ranging from 1 to 5). OVP and Youtube both contain 50 videos, whose annotations are sets of key-frames, produced by 5 users. The video duration ranges from 1 to 4 minutes for OVP, and from 1 to 10 minutes for Youtube. Both datasets are comprised of videos with diverse video content, such as documentaries, educational, ephemeral, historical, and lecture videos (OVP dataset) and cartoons, news, sports, commercials, TV-shows and home videos (Youtube dataset). Given the size of each of these datasets, we argue that there is a lack of large-scale annotated datasets that could be useful for improving the training of complex supervised deep learning architectures.

Some less-commonly used datasets for video summarization are CoSum \cite{7298981}, MED-summaries \cite{10.1007/978-3-319-10599-4_35}, Video Titles in the Wild (VTW) \cite{10.1007/978-3-319-46475-6_38}, League of Legends (LoL) \cite{fu-etal-2017-video} and 
FVPSum \cite{10.1007/978-3-030-01267-0_5}. CoSum has been created to evaluate video co-summarization. It consists of $51$ videos that were downloaded from YouTube using $10$ query terms that relate to the video content of the SumMe dataset. Each video is approximately $4$ minutes long and it is annotated by $3$ different users, who have selected sets of key-fragments. The MED-summaries dataset contains $160$ annotated videos from the TRECVID $2011$ MED dataset. $60$ videos form the validation set (from $15$ event categories) and the remaining $100$ videos form the test set (from $10$ event categories), with most of them being $1$ to $5$ minutes long. The annotations come as one set of importance scores, averaged over $1$ to $4$ annotators. As far as the VTW dataset is concerned, it includes $18100$ open domain videos, with $2000$ of them being annotated in the form of sub-shot level highlight scores. The videos are user-generated, untrimmed videos that contain a highlight event and have an average duration of $1.5$ minutes. Regarding LoL, it has $218$ long videos ($30$ to $50$ minutes), displaying game matches from the North American League of Legends Championship Series (NALCS). The annotations derive from a Youtube channel that provides community-generated highlights (videos with duration $5$ to $7$ minutes). Therefore, one set of key-fragments is available for each video. Finally, FPVSum is a first-person video summarization dataset. It contains $98$ videos (more than $7$ hours total duration) from $14$ categories of GoPro viewer-friendly videos. For each category, about $35\%$ of the video sequences are annotated with ground-truth scores by at least $10$ users, while the remaining are viewed as the unlabeled examples. The main characteristics of each of the above discussed datasets, are briefly presented in Table \ref{tab:datasets}.

It is worth mentioning that in this work we focus only on datasets that are fit for evaluating video summarization methods, namely datasets that contain ground truth annotations regarding the summary or the frame/fragment-level importance of each video. Other datasets might be also used by some works for network pre-training purposes, but they do not concern the present work. Table~\ref{tab:methods_n_datasets} summarizes the datasets utilized by the deep-learning-based approaches for video summarization. From this table it is clear that the SumMe and TVSum datasets are, by far, the most commonly used ones. OVP and Youtube are also utilized in a few works, but mainly for data augmentation purposes.

\begin{table*}[]
\begin{center}
\begin{tabular}{|c|c|c|c|c|c|}
\hline
Dataset        & \# of videos & duration (min) & content                                                                                                                    & type of annotations             & \# of annotators per video \\ \hline
SumMe~\cite{10.1007/978-3-319-10584-0_33}   & 25       & 1 - 6          & holidays, events, sports  & multiple sets of key-fragments  & 15 - 18                \\ \hline
TVSum~\cite{7299154}   & 50       & 2 - 10         & \begin{tabular}[c]{@{}c@{}}news, how-to's, user-generated, \\ documentaries\\ (10 categories - 5 videos each)\end{tabular} & multiple fragment-level scores  & 20                     \\ \hline
OVP~\cite{10.1016/j.patrec.2010.08.004}     & 50       & 1 - 4          & \begin{tabular}[c]{@{}c@{}}documentary, educational,\\  ephemeral, historical, lecture\end{tabular}                        & multiple sets of key-frames     & 5                      \\ \hline
Youtube~\cite{10.1016/j.patrec.2010.08.004} & 50       & 1 - 10         & \begin{tabular}[c]{@{}c@{}}cartoons, sports, tv-shows, \\ commercial, home videos\end{tabular}                             & multiple sets of key-frames     & 5                      \\ \hline
CoSum~\cite{7298981}   & 51       & $\sim$ 4        & \begin{tabular}[c]{@{}c@{}}holidays, events, sports\\ (10 categories)\end{tabular}                                         & multiple sets of key-fragments  & 3                      \\ \hline
MED~\cite{10.1007/978-3-319-10599-4_35}     & 160      & 1 - 5          & 15 categories of various genres                                                                                            & one set of imp. scores          & 1 - 4                  \\ \hline
VTW~\cite{10.1007/978-3-319-46475-6_38}     & 2000     & 1.5 (avg)      & \begin{tabular}[c]{@{}c@{}}user-generated videos that \\ contain a highlight event\end{tabular}                            & sub-shot level highlight scores & -                      \\ \hline
LoL~\cite{fu-etal-2017-video}     & 218      & 30 - 50        & \begin{tabular}[c]{@{}c@{}}matches from a League \\ of Legends tournament\end{tabular}                                     & one set of key-fragments        & 1                      \\ \hline
FPVSum~\cite{10.1007/978-3-030-01267-0_5}     & 98      & 4.3 (avg)    &    \begin{tabular}[c]{@{}c@{}}first-person videos \\ (14 categories)\end{tabular}                                     & multiple frame-level scores       & 10                      \\ \hline
\end{tabular}
\end{center}
\caption{Datasets for video summarization and their main characteristics.}
\label{tab:datasets}
\end{table*}

\begin{table*}[]
\begin{center}
\begin{tabular}{|l|c|c|c|c|c|c|c|c|c|c|}
\hline
Method           & SumMe                      & TVSum                      & OVP                        & Youtube                    & CoSum                      & MED                        & VTW                        & LoL            & FPVSum                                 \\ \hline
vsLSTM (2016)~\cite{10.1007/978-3-319-46478-7_47}  & \ding{51} & \ding{51} & \ding{51} & \ding{51} &                            &                            &                            &                            &                         \\ \hline
dppLSTM (2016)~\cite{10.1007/978-3-319-46478-7_47}  & \ding{51} & \ding{51} & \ding{51} & \ding{51} &                            &                            &                            &                            &                         \\ \hline
H-RNN (2017)~\cite{Zhao:2017:HRN:3123266.3123328}            & \ding{51} & \ding{51} &                            &                            &                            & \ding{51} & \ding{51} &                            &                         \\ \hline
SUM-GAN (2017)~\cite{Mahasseni2017UnsupervisedVS}          & \ding{51} & \ding{51} & \ding{51} & \ding{51} &                            &                            &                            &                            &                         \\ \hline
DeSumNet (2017)~\cite{8237657}         &                            & \ding{51} &                            &                            & \ding{51} &                            &                            &                            &                          \\ \hline
VS-DSF (2017)~\cite{10.1007/978-3-319-54193-8_23}           & \ding{51} &                            &                            &                            &                            &                            &                            &                            &                         \\ \hline
HSA-RNN (2018)~\cite{ZhaoHSARNNHS}          & \ding{51} & \ding{51} &                            &                            & \ding{51} &                            & \ding{51} &                            &                         \\ \hline
SUM-FCN (2018)~\cite{10.1007/978-3-030-01258-8_22}          & \ding{51} & \ding{51} & \ding{51} & \ding{51} &                            &                            &                            &                            &                         \\ \hline
MAVS (2018)~\cite{Feng:2018:EVS:3240508.3240651}             & \ding{51} & \ding{51} &                            &                            &                            &                            &                            &                            &                       \\ \hline
DR-DSN (2018)~\cite{Zhou2018DeepRL}           & \ding{51} & \ding{51} & \ding{51} & \ding{51} &                            &                            &                            &                            &                       \\ \hline
Online Motion-AE (2018)~\cite{ZHANG2018}  & \ding{51}  & \ding{51} &  &  \ding{51}  &  &  &   &   &  \\ \hline
FPVSF (2018)~\cite{10.1007/978-3-030-01267-0_5}          & \ding{51} & \ding{51} &  &  &                            &                            &                            &                            &        \ding{51}                 \\ \hline
VESD (2018)~\cite{10.1007/978-3-030-01264-9_12}          &  & \ding{51} &  &  &       \ding{51}                     &                            &                            &                            &                       \\ \hline
DQSN (2018)~\cite{Zhou2018VideoSB}             &                            & \ding{51} &                            &                            & \ding{51} &                            &                            &                            &                     \\ \hline
SA-SUM (2018)~\cite{Wei2018VideoSV}           & \ding{51} & \ding{51} &                            & \ding{51} &                            &                            &                            &                            &                      \\ \hline
vsLSTM+Att (2019)~\cite{10.1007/978-3-030-05716-9_6}  & \ding{51}  & \ding{51} & \ding{51} & \ding{51}  &  &  &   &   &  \\ \hline
dppLSTM+Att (2019)~\cite{10.1007/978-3-030-05716-9_6}  & \ding{51}  & \ding{51} & \ding{51} & \ding{51}  &  &  &   &   &  \\ \hline
VASNet (2019)~\cite{10.1007/978-3-030-21074-8_4}           & \ding{51} & \ding{51} & \ding{51} & \ding{51} &                            &                            &                            &                            &                          \\ \hline
H-MAN (2019)~\cite{liu2019learning}  & \ding{51}  & \ding{51} & \ding{51} & \ding{51}  &  &  &   &   & \\ \hline
SMN (2019)~\cite{Wang:2019:SMN:3343031.3350992}       & \ding{51} & \ding{51} & \ding{51} & \ding{51} &  &  &  &  &   \\ \hline
CRSum (2019)~\cite{yuan2019spatiotemporal}  & \ding{51}  & \ding{51} &  &   &  &  & \ding{51}  &   &  \\ \hline
SMLD (2019)~\cite{chu2019spatiotemporal}       & \ding{51} & \ding{51} &                            &                            &                            &                            &                            &                            &                        \\ \hline
ActionRanking (2019)~\cite{8659061}  & \ding{51}  & \ding{51} & \ding{51} & \ding{51}  &  &  &   &   &  \\ \hline
DTR-GAN (2019)~\cite{Zhang:2019:DDT:3321408.3322622}          &  & \ding{51} &                            &                            &                            &                            &                            &                            &                      \\ \hline
Ptr-Net (2019)~\cite{DBLP:conf/wacv/FuTC19}          & \ding{51} & \ding{51} &                            & \ding{51} &                            &                            &                            & \ding{51} &                           \\ \hline
SUM-GAN-sl (2019)~\cite{Apostolidis:2019:SLA:3347449.3357482}       & \ding{51} & \ding{51} &                            &                            &                            &                            &                            &                            &                         \\ \hline
CSNet (2019)~\cite{jung2019discriminative}             & \ding{51}                      & \ding{51} &  \ding{51}                  & \ding{51}               &  &                            &                            &                            &                        \\ \hline
Cycle-SUM (2019)~\cite{Yuan2019CycleSUMCA}        & \ding{51} & \ding{51} &                            &                            &                            &                            &                            &                            &                        \\ \hline
ACGAN (2019)~\cite{He:2019:UVS:3343031.3351056}       & \ding{51} & \ding{51} & \ding{51} & \ding{51} &  &  &  &  &  \\ \hline
UnpairedVSN (2019)~\cite{Rochan_2019_CVPR}       & \ding{51} & \ding{51} & \ding{51} & \ding{51} &  &  &  &  &    \\ \hline
EDSN (2019)~\cite{gonuguntla2019enhanced}       & \ding{51} & \ding{51} &                            &                            &                            &                            &                            &                            &                         \\ \hline
WS-HRL (2019)~\cite{10.1145/3338533.3366583}  & \ding{51}  & \ding{51} & \ding{51} & \ding{51}  &  &  &   &   &  \\ \hline
DSSE (2019)~\cite{8101557}             &                            & \ding{51} &                            &                            &                            &                            &                            &                            &                         \\ \hline
A-AVS (2020)~\cite{8667390}     & \ding{51} & \ding{51} & \ding{51} & \ding{51} &                            &                            &                            &                            &                        \\ \hline
M-AVS (2020)~\cite{8667390}     & \ding{51} & \ding{51} & \ding{51} & \ding{51} &                            &                            &                            &                            &                        \\ \hline
DASP (2020)~\cite{JI2020200}      & \ding{51} & \ding{51} &                            &      \ding{51}                      &                            &                            &                            &                            &                        \\ \hline
SF-CVS (2020)~\cite{8601376}  & \ding{51}  & \ding{51} &  & \ding{51}  &  &  &   &   &  \\ \hline
SUM-GAN-AAE (2020)~\cite{10.1007/978-3-030-37731-1_40}      & \ding{51} & \ding{51} &                            &                            &                            &                            &                            &                            &                      \\ \hline
PCDL (2020)~\cite{8924889}  & \ding{51}  & \ding{51} & \ding{51} & \ding{51}  &  &  &   &   &  \\ \hline
AC-SUM-GAN (2020)~\cite{9259058}      & \ding{51} & \ding{51} &                            &                            &                            &                            &                            &                            &                      \\ \hline
TTH-RNN (2020)~\cite{9037206}            & \ding{51} & \ding{51} &                            &                            &                            & \ding{51} & \ding{51} &                            &                         \\ \hline
GL-RPE (2020)~\cite{10.1007/978-3-030-58595-2_11}      & \ding{51} & \ding{51} &                            &                            &                            &                            &                            &                            &                      \\ \hline
SUM-$Ind_{LU}$ (2021)~\cite{DBLP:journals/mta/YalinizI21}     & \ding{51} & \ding{51} &                            &                            &                            &                            &                            &                            &                      \\ \hline
SUM-GDA (2021)~\cite{LI2021107677}      & \ding{51} & \ding{51} & \ding{51} & \ding{51}   &        &        &    \ding{51}  &       &      \\ \hline
\end{tabular}
\end{center}
\caption{Datasets used by each deep-learning-based method for evaluating video summarization performance.}
\label{tab:methods_n_datasets}
\end{table*}

\subsection{Evaluation protocols and measures}
Several approaches have been proposed in the literature for evaluating the performance of video summarization. A categorization of these approaches, along with a brief presentation of their main characteristics, is provided in Table \ref{tab:eval_prot}. In the sequel we discuss in more details these evaluation protocols in chronological order, to show the evolution of ideas on the assessment of video summarization methods.

\begin{table*}[]
\begin{center}
\begin{tabular}{|l|l|}
\hline
\multicolumn{2}{|c|}{\textbf{Qualitative evaluation of video storyboards based on user studies}}                                                                                                                                                                                                                                                                                                                                                                                                                                                                                                                                                                                                                                                                                                                                                                                                                                                                                \\ \hline
\textbf{\begin{tabular}[c]{@{}l@{}}Relevant works\end{tabular}}                                                                                                                                                                                                                                                                                                                                                                                                                                                                                                                                                                                     & \textbf{Adopted criteria}                                                                                                                                                                                                                               \\ \hline
Avila et al. (2008) \cite{4654149}                                                                                                                                                                                                                                                                                                                                                                                                                                                                                                                                                                                                                                        & \begin{tabular}[c]{@{}l@{}}Relevance of each key-frame with the video content and redundant or missing information in the\\ key-frame set\end{tabular}                                                                                     \\ \hline
Ejaz et al. (2014) \cite{EJAZ2014993}                                                                                                                                                                                                                                                                                                                                                                                                                                                                                                                                                                                                                                     & Informativeness and enjoyability of the key-frame set                                                                                                                                                                                                   \\ \hline \hline
\multicolumn{2}{|c|}{\textbf{Quantitative evaluation of video storyboards based on ground-truth annotations}}    \\ \hline
\textbf{\begin{tabular}[c]{@{}l@{}}Relevant works\end{tabular}}                                                                                                                                                                                                                                                                                                                                                                                                                                                                                                                                                                       & \textbf{Used measures}                                                                                                                                                                                                                                  \\ \hline
Chasanis et al. (2008) \cite{10.1007/978-3-540-87536-9_87}                                                                                                                                                                                                                                                                                                                                                                                                                                                                                                                                                                                                               & \begin{tabular}[c]{@{}l@{}}Fidelity (min. distance of a frame from the key-frame set) and Shot Reconstruction Degree (level\\ of reconstruction of the entire frame sequence using the key-frame set)\end{tabular}                                    \\ \hline
\begin{tabular}[c]{@{}l@{}}Avila et al. (2011) \cite{10.1016/j.patrec.2010.08.004}\\ (also used in \cite{10.1016/j.jvcir.2012.06.013,10.1016/j.patrec.2011.08.007,6656190,10.1007/s10844-016-0441-4})\end{tabular}                                                                                                                                                                                                                                                                                                                                                                                                                                       & \begin{tabular}[c]{@{}l@{}}Overlap between machine- and multiple user-generated key-frame-based summaries measured by\\ Accuracy and Error Rates (known as "Comparison of User Summaries")\end{tabular}                                              \\ \hline
\begin{tabular}[c]{@{}l@{}}Mahmoud et al. (2013) \cite{6786125}\\ (also used in \cite{NIPS2014_5413,10.1145/2632267,MEI2015522,7406489})\end{tabular}                                                                                                                                                                                                                                                                                                                                                                                                                                                                                                   & \begin{tabular}[c]{@{}l@{}}Overlap between machine- and multiple user-generated key-frame-based summaries measured by\\ Precision, Recall, F-Score\end{tabular}                                                                                         \\ \hline \hline
\multicolumn{2}{|c|}{\textbf{Quantitative evaluation of video skims based on ground-truth annotations}} \\ \hline
\textbf{\begin{tabular}[c]{@{}l@{}}Relevant works\end{tabular}}  & \textbf{Used measures} \\ \hline
\begin{tabular}[c]{@{}l@{}}Gygli et al. (2014) \cite{10.1007/978-3-319-10584-0_33}\\ (also used in \cite{7298928,10.1007/978-3-319-46478-7_47,Zhao:2017:HRN:3123266.3123328,10.1007/978-3-030-21074-8_4,8667390,Wang:2019:SMN:3343031.3350992,8659061}\\ \cite{Apostolidis:2019:SLA:3347449.3357482,jung2019discriminative,10.1007/978-3-030-37731-1_40,He:2019:UVS:3343031.3351056,Rochan_2019_CVPR,Zhou2018DeepRL,8924889,Zhou2018VideoSB,9259058,10.1007/978-3-030-58595-2_11,LI2021107677,9037206,lal2019online}\\ \cite{chu2019spatiotemporal,7299154,ZhaoHSARNNHS,liu2019learning,10.1007/978-3-319-54193-8_23,Wei2018VideoSV,8245827,7904630})\end{tabular} & \begin{tabular}[c]{@{}l@{}}Overlap between machine- and multiple user-generated key-fragment-based summaries measured\\ by Precision, Recall, F-Score\end{tabular}                                                                                      \\ \hline
\begin{tabular}[c]{@{}l@{}}Mahasseni et al. (2017) \cite{Mahasseni2017UnsupervisedVS}\\ (also used in \cite{Yuan2019CycleSUMCA, Zhang:2019:DDT:3321408.3322622, DBLP:conf/wacv/FuTC19, DBLP:journals/mta/YalinizI21})\end{tabular}                                                                                                                                                                                                                                                                                                                                                                                                                                                                                                                & \begin{tabular}[c]{@{}l@{}}Overlap between machine- and single ground-truth key-fragment-based summary measured by\\ Precision, Recall, F-Score\end{tabular}                                                                                            \\ \hline
\begin{tabular}[c]{@{}l@{}}Otani et al. (2019) \cite{otani2018vsumeval}\\ (also used in \cite{10.1145/3338533.3366583,10.1007/978-3-030-58595-2_11})\end{tabular}                                                                                                                                                                                                                                                                                                                                                                                                                                                                                       & \begin{tabular}[c]{@{}l@{}}Alignment between machine- and multiple user-generated series of frame-level importance scores\\ using the Kendall and Spearman rank correlation coefficients\end{tabular}                                                \\ \hline
Apostolidis et al. (2020) \cite{10.1145/3394171.3413632}                                                                                                                                                                                                                                                                                                                                                                                                                                                                                                                                                                                                                  & \begin{tabular}[c]{@{}l@{}}Extension of the one from Gygli et al. \cite{10.1007/978-3-319-10584-0_33}. The performance of the machine-based summarizer\\ is divided by the performance of a random summarizer (known as "Performance Over Random")\end{tabular} \\ \hline
\end{tabular}
\end{center}
\caption{Proposed protocols for the evaluation of video summarization methods.}
\label{tab:eval_prot}
\end{table*}

\subsubsection{Evaluating video storyboards}
Early video summarization techniques created a static summary of the video content with the help of representative key-frames. First attempts towards the evaluation of the created key-frame-based summaries were based on user studies that made use of human judges for evaluating the resulting quality of the summaries. Judgment was based on specific criteria, such as the relevance of each key-frame with the video content, redundant or missing information \cite{4654149}, or the informativeness and enjoyability of the summary \cite{EJAZ2014993}. Although this can be a useful way to evaluate results, the procedure of evaluating each resulting summary by users can be time-consuming and the evaluation results can not be easily reproduced or used in future comparisons. To overcome the deficiencies of user studies, other works evaluate their key-frame-based summaries using objective measures and ground-truth summaries. In this context, Chasanis et al. (2008) \cite{10.1007/978-3-540-87536-9_87} estimated the quality of the produced summaries using the Fidelity measure \cite{809161} and the Shot Reconstruction Degree criterion \cite{Liu2004ShotRD}. A different approach, termed ``Comparison of User Summaries'', was proposed in \cite{10.1016/j.patrec.2010.08.004} and evaluates the generated summary according to its overlap with predefined key-frame-based user summaries. Comparison is performed at a key-frame-basis and the quality of the generated summary is quantified by computing the Accuracy and Error Rates based on the number of matched and non-matched key-frames respectively. This approach was also used in~\cite{10.1016/j.jvcir.2012.06.013,10.1016/j.patrec.2011.08.007,6656190,10.1007/s10844-016-0441-4}. A similar methodology was followed in \cite{6786125}. Instead of Accuracy and Error Rates, in \cite{6786125} the evaluation relies on the well-known Precision, Recall and F-Score measures. This protocol was also used in the supervised video summarization approach of~\cite{NIPS2014_5413}, while a variation of it was used in \cite{10.1145/2632267,MEI2015522,7406489}. In the latter case, additionally to their visual similarity, two frames are considered a match only if they are temporally no more than a specific number of frames apart.

\subsubsection{Evaluating video skims}
\label{subsubsec:eval_prot}
Most recent algorithms tackle video summarization by creating a dynamic video summary (video skim). For this, they select the most representative video fragments and join them in a sequence to form a shorter video. The evaluation methodologies of these works assess the quality of video skims according to their alignment with human preferences. Contrary to the early approaches for video storyboard generation that utilized qualitative evaluation methodologies, these works extend the evaluation protocols of the late video storyboard generation approaches, and perform the evaluation using ground-truth data and objective measures. A first attempt was made in \cite{10.1007/978-3-319-10584-0_33}, where an evaluation approach along with the SumMe dataset for video summarization were introduced. According to this approach, the videos are first segmented into consecutive and non-overlapping fragments in order to enable matching between key-fragment-based summaries (i.e., to compare the user-generated with the automatically-created summary). Then, based on the scores computed by a video summarization algorithm for the fragments of a given video, an optimal subset of them (key-fragments) is selected and forms the summary. The alignment of this summary with the user summaries for this video is evaluated by computing F-Score in a pairwise manner. In particular, the F-Score for the summary of the $i^{th}$ video is computed as follows:
\begin{equation}
    F_i=\frac{1}{N_i}\sum_{j=1}^{N_i}2\;\frac{P_{i,j}\;R_{i,j}}{P_{i,j}+R_{i,j}}
\end{equation}
where $N_{i}$ is the number of available user-generated summaries for the $i^{th}$ test video, $P_{i,j}$ and $R_{i,j}$ are the Precision and Recall against the $j^{th}$ user summary, and they are both computed on a per-frame basis. This methodology was adopted also in~\cite{7298928, 10.1007/978-3-319-54193-8_23} and~\cite{7299154}. The latter work introduced another dataset, called TVSum. As in \cite{10.1007/978-3-319-10584-0_33}, the shots of the videos of the TVSum dataset were defined through automatic video segmentation. Based on the results of a video summarization algorithm, the computed (frame- or fragment-level) scores are used to define the sequence of selected video fragments and produce the summary. Similarly to \cite{10.1007/978-3-319-10584-0_33}, for a given video of the TVSum dataset, the agreement of the created summary with the user summaries is quantified by F-Score.

The evaluation approach and benchmark datasets of \cite{10.1007/978-3-319-10584-0_33} and \cite{7299154} were jointly used to evaluate the summarization performance in \cite{10.1007/978-3-319-46478-7_47}. After defining a new segmentation for the videos of both datasets, Zhang et al. \cite{10.1007/978-3-319-46478-7_47} evaluated the efficiency of their method on both datasets based on the multiple user-generated summaries for each video. Moreover, they documented the needed conversions from frame-level importance scores to key-fragment-based summaries in the Supplementary Material of~\cite{10.1007/978-3-319-46478-7_47}. The typical settings about the data split into training and testing ($80\%$ for training and $20\%$ for testing) and the target summary length ($\leq 15\%$ of the video duration) were used, and the evaluation was based on F-Score. Experiments were conducted five times and the authors report the average performance and the standard deviation (STD). The above described evaluation protocol - with slight variations that relate to the number of iterations using different randomly created splits of the data (5-splits; 10-splits; ``few''-splits; 5-fold cross validation), the way that the computed F-Scores from the pairwise comparisons with the different user summaries are taken under consideration (maximum value is kept for SumMe according to~\cite{7298928}; average value is kept for TVSum) to form the F-Score for a given test video, and the way the average performance of these multiple runs is indicated (mean of highest performance for each run; best mean performance at the same training epoch for all runs) - has been adopted by the vast majority of the state-of-the-art works on video summarization (see~\cite{8667390,8245827,Zhao:2017:HRN:3123266.3123328,Zhou2018DeepRL,Wei2018VideoSV,Zhou2018VideoSB,ZhaoHSARNNHS,10.1007/978-3-030-21074-8_4,8659061,jung2019discriminative,Apostolidis:2019:SLA:3347449.3357482,10.1007/978-3-030-37731-1_40,8924889,chu2019spatiotemporal,He:2019:UVS:3343031.3351056,Rochan_2019_CVPR,Wang:2019:SMN:3343031.3350992,lal2019online,7904630,liu2019learning,9259058,9037206,10.1007/978-3-030-58595-2_11,LI2021107677}). Hence, it can be seen as the currently established approach for assessing the performance of video summarization algorithms.

A slightly different evaluation approach calculates the agreement with a single ground-truth summary, instead of multiple user summaries. This single ground-truth summary is computed by averaging the key-fragment summaries per frame, in the case of multiple key-fragment-based user summaries and by averaging all users' scores for a video on a frame-basis, in the case of frame-level importance scores. This approach is utilized by only a few methods (i.e., \cite{Mahasseni2017UnsupervisedVS, Yuan2019CycleSUMCA, Zhang:2019:DDT:3321408.3322622, DBLP:conf/wacv/FuTC19, DBLP:journals/mta/YalinizI21}), as it doesn't maintain the original opinion of every user, leading to a less firm evaluation. 

Another evaluation approach was proposed in~\cite{otani2018vsumeval}. This method is independent of any predefined fragmentation of the video. The user-generated frame-level importance scores for the TVSum videos are considered as rankings, and two rank correlation coefficients, namely Kendall $\tau$~\cite{kendall1945treatment} and Spearman $\rho$~\cite{kokoska2000crc} coefficients, are used to evaluate the summary. However, these measures can be used only on datasets that follow the TVSum annotations and methods that produce the same type of results (i.e., frame-level importance scores). This methodology was used (in addition to the established protocol) to evaluate the algorithms in \cite{10.1145/3338533.3366583,10.1007/978-3-030-58595-2_11}.

Last but not least, a new evaluation protocol for video summarization was presented in \cite{10.1145/3394171.3413632}. This work started by evaluating the performance of five publicly-available methods under a large-scale experimental setting with $50$ randomly-created data splits of the SumMe and TVSum datasets, where the performance is evaluated using F-Score. The conducted study showed that the results reported in the relevant papers are not always congruent with the performance on the large-scale experiment, and that the F-Score is not most suitable for comparing algorithms that were run on different splits. For this, Apostolidis et al. (2020) \cite{10.1145/3394171.3413632} proposed a new evaluation protocol, called ``Performance over Random'', that estimates the difficulty of each used data split and utilizes this information during the evaluation process.

\section{Performance comparisons}
\label{sec:perf_comp}

\subsection{Quantitative comparisons}

In this section we present the performance of the reviewed deep-learning-based video summarization approaches on the SumMe and TVSum datasets, as reported in the corresponding papers. We focus only on these two datasets, since they are prevalent in the relevant literature.

Table \ref{tab:sup_comp} reports the performance of (weakly-)supervised video summarization algorithms that have been assessed via the established evaluation approach (i.e., using the entire set of available user summaries for a given video). In the same table, we report the performance of a random summarizer. To estimate this performance, importance scores are randomly assigned to the frames of a given video based on a uniform distribution of probabilities. The corresponding fragment-level scores are then used to form video summaries using the Knapsack algorithm and a length budget of maximum $15\%$ of the original video's duration. Random summarization is performed $100$ times for each video, and the overall average score is reported (for further details please check the relevant algorithm in \cite{10.1145/3394171.3413632}). Moreover, the rightmost column of this table provides details about the number and type of data splits that were used for the evaluation, with ''X Rand'' denoting X randomly-created splits (with X being equal to $1$, $5$, $10$ or an unspecified value (this case is noted as M Rand)) and ''5 FCV'' denoting 5-fold cross validation. Finally, the algorithms are listed according to their average ranking on both datasets (see the $4^{th}$ column ``Avg Rnk'').

\begin{table}[t]
\begin{center}
\resizebox{\columnwidth}{!}{%
\begin{tabular}{|l|ll|ll|l|c|}
\hline
 & \multicolumn{2}{c|}{SumMe} & \multicolumn{2}{c|}{TVSum} & Avg & Data \\
\cline{2-5}
 & F1 & Rnk & F1 & Rnk  &  Rnk & splits  \\ \hline
Random summary		                                  & 40.2	      & 27	   & 54.4           & 24     &  25.5 & $-$  \\ \hline
vsLSTM~\cite{10.1007/978-3-319-46478-7_47}            & 37.6          & 30     & 54.2           &  25    &  27.5 &  1 Rand  \\
dppLSTM~\cite{10.1007/978-3-319-46478-7_47}           & 38.6          & 29     & 54.7           &  23    &  26 &  1 Rand  \\
SASUM~\cite{Wei2018VideoSV}    & 40.6  & 25   & 53.9 & 26  & 25.5 & 10 Rand  \\
ActionRanking~\cite{8659061}                          & 40.1          &  28    & 56.3           & 21     &  24.5 & 1 Rand \\
$\diamond$FPVSF~\cite{10.1007/978-3-030-01267-0_5}            & 41.9          & 24     & $-$~\tablefootnote{In this work the TVSum dataset is used as the third-person labeled data for training purposes only, so we do not present any result here.}           & $-$    &  24  & $-$ \\
vsLSTM+Att~\cite{10.1007/978-3-030-05716-9_6}   & 43.2  & 22  & $-$~\tablefootnote{\label{note1}The authors of this literature work evaluate their method on TVSum using a protocol that differs from the typical protocol used with this dataset so, we do not present this result here.}  & $-$  & 22 & 1 Rand \\
H-RNN~\cite{Zhao:2017:HRN:3123266.3123328}            & 42.1          & 23     & 57.9           & 18    &  20.5  & $-$ \\
DR-DSN$_{sup}$~\cite{Zhou2018DeepRL}                  & 42.1          &  23     & 58.1           & 16     &  19.5  & 5 FCV  \\
dppLSTM+Att~\cite{10.1007/978-3-030-05716-9_6}  & 43.8  & 19  & $-$~\textsuperscript{\ref{note1}}  & $-$  & 19  & 1 Rand \\ 
$+$DSSE~\cite{8101557}                                   & $-$             & $-$     & 57.0           & 19     &  19  & $-$   \\
$\diamond$WS-HRL~\cite{10.1145/3338533.3366583}    & 43.6  & 21   & 58.4 & 14  & 17.5  & 5 FCV \\
PCDL$_{sup}$~\cite{8924889}    & 43.7  & 20   & 59.2 & 10  & 15  & 1 Rand \\
SF-CVS~\cite{8601376}   & 46.0  & 12  & 58.0  & 17  & 14.5  & $-$ \\
SASUM$_{fullysup}$~\cite{Wei2018VideoSV}    & 45.3  & 14   & 58.2 & 15  & 14.5   & 10 Rand \\
UnpairedVSN$_{psup}$~\cite{Rochan_2019_CVPR}          & 48.0          & 7     & 56.1           &  22    &  14.5  & 5 Rand \\
SUM-FCN~\cite{10.1007/978-3-030-01258-8_22}           & 47.5          & 9     & 56.8           & 20     &  14.5  & M Rand \\
MAVS~\cite{Feng:2018:EVS:3240508.3240651}             & 40.3          & 26     & \textbf{66.8}  &  1     &  13.5  & 5 FCV  \\
A-AVS~\cite{8667390}      & 43.9 & 18  & 59.4 & 9  & 13.5 & 5 Rand \\
CRSum~\cite{yuan2019spatiotemporal}    & 47.3  & 10    & 58.0 & 17  & 13.5 &  5 FCV  \\
HSA-RNN~\cite{ZhaoHSARNNHS}		                      & 44.1          & 17      & 59.8           & 8      &  12.5  &  $-$ \\
$+$DQSN~\cite{Zhou2018VideoSB}                           & $-$             & $-$      & 58.6       & 12      &  12  & 5 FCV \\
TTH-RNN~\cite{9037206}            & 44.3          & 16     & 60.2           & 7    &  11.5  & $-$ \\
M-AVS~\cite{8667390}      & 44.4 & 15  & 61.0  &  5  & 10 & 5 Rand \\ 
ACGAN$_{sup}$~\cite{He:2019:UVS:3343031.3351056}	  & 47.2	      &  11     & 59.4           & 9     &  10  & 5 FCV  \\
SUM-DeepLab~\cite{10.1007/978-3-030-01258-8_22}       & 48.8          & 5      & 58.4           & 14     &  9.5  & M Rand  \\
CSNet$_{sup}$~\cite{jung2019discriminative}		      & 48.6	      &	6      & 58.5           &  13     &  9.5  & 5 FCV \\
DASP~\cite{JI2020200}  & 45.5  &  13  & 63.6 & 3  & 8   &  5 Rand \\
SMLD~\cite{chu2019spatiotemporal}    & 47.6  &  8  & 61.0 & 5  & 6.5   &  5 FCV \\
SUM-GDA~\cite{LI2021107677}           & 52.8          &  2     & 58.9           & 11      &  6.5  & 5 FCV \\
H-MAN~\cite{liu2019learning}    & 51.8  & 3   & 60.4 & 6  & 4.5  &  5 FCV \\
VASNet~\cite{10.1007/978-3-030-21074-8_4}             & 49.7          &  4     & 61.4           & 4      &  4  & 5 FCV \\
SMN~\cite{Wang:2019:SMN:3343031.3350992}	          & \textbf{58.3} & 1      & 64.5           & 2     &  \textbf{1.5} &  5 Rand \\ \hline
\end{tabular}}
\end{center}
\caption{Comparison (F1: F-Score (\%)) of supervised and weakly-supervised video summarization approaches on SumMe and TVSum. Weakly-supervised methods marked with $\diamond$. Multimodal approaches marked with $+$}
\label{tab:sup_comp}
\end{table}

Based on the reported performances, we can make the following observations:
\begin{itemize}
    \item The best-performing supervised approaches utilize tailored attention mechanisms (VASNet, H-MAN, SUM-GDA, DASP, CSNet$_{sup}$) or memory networks (SMN) to capture variable- and long-range temporal dependencies.
    \item Some works (e.g., MAVS, DASP, M-AVS, A-AVS, TTH-RNN) exhibit high performance in one of the datasets and very low or even random performance in the other dataset. This poorly-balanced performance indicates techniques that may be highly-adapted to a specific dataset.
    \item The use of additional modalities of the video (mainly the associated text-based video metadata) does not seem to help, as the multimodal methods (DSSE, DQSN) are not competitive compared to the unimodal ones that rely on the analysis of the visual content only.
    \item The use of weak labels instead of a full set of human annotations does not enable good summarization, as the weakly-supervised methods (FPVSF, WS-HRL) perform similarly to the random summarizer on SumMe and poorly on TVSum.
    \item Finally, a few methods (placed at the top of Table \ref{tab:sup_comp}) show random performance in at least one of the used datasets.
\end{itemize}

Table \ref{tab:unsup_comp} presents the performance of unsupervised video summarization methods that have been assessed with the same evaluation approach (i.e., using the entire set of available user summaries for a given video). As in Table \ref{tab:sup_comp}, this table also reports the performance of a random summarizer and provides details about the number and type of data splits that were used for the evaluation (see the rightmost column). Once again, the algorithms are presented according to their average ranking on both datasets (see the $4^{th}$ column ``Avg Rnk'').

Based on the reported results, we can make the following remarks:
\begin{itemize}
    \item The use of GANs for learning summarization in a fully-unsupervised manner is a good choice, as most of the best-performing methods (AC-SUM-GAN, CSNet, CSNet+GL+RPE, SUM-GAN-AAE, SUM-GAN-sl) rely on this framework.
    \item The use of attention mechanisms helps to identify the important parts of the video, as a few of the best-performing algorithms (CSNet, CSNet+GL+RPE, SUM-GDA$_{unsup}$, SUM-GAN-AAE) utilize such mechanisms. The benefits of using such a mechanism are also documented through the comparison of the SUM-GAN-sl and SUM-GAN-AAE techniques. The replacement of the Variational Auto-Encoder (that is used in SUM-GAN-sl) by a deterministic Attention Auto-Encoder (which is introduced in SUM-GAN-AAE) results in a clear performance improvement on the SumMe dataset, while maintaining the same levels of summarization performance on TVSum.
    \item Techniques that rely on reward functions and reinforcement learning (DR-DSN, EDSN) are not so competitive compared to GAN-based methods, especially on SumMe.
    \item Finally, a few methods (placed at the top of Table \ref{tab:unsup_comp}) perform approximately equally to the random summarizer.
\end{itemize}

\begin{table}
\begin{center}
\resizebox{\columnwidth}{!}{%
\begin{tabular}{|l|ll|ll|l|c|}
\hline
 & \multicolumn{2}{c|}{SumMe} & \multicolumn{2}{c|}{TVSum} & Avg & Data\\
\cline{2-5}                                          
   & F1          & Rnk    & F1           & Rnk    & Rnk & splits\\ \hline
Random summary	 & 40.2	       & 13	    & 54.4           & 11      & 12 & $-$ \\ \hline
Online Motion-AE~\cite{ZHANG2018}  & 37.7          & 14      & 51.5           & 13      & 13.5 &  $-$ \\ 
SUM-FCN$_{unsup}$~\cite{10.1007/978-3-030-01258-8_22} & 41.5  & 11   & 52.7 & 12  & 11.5 & M Rand \\
DR-DSN~\cite{Zhou2018DeepRL}    & 41.4          & 12      & 57.6           & 8      & 10 & 5 FCV  \\
EDSN~\cite{gonuguntla2019enhanced}    & 42.6  & 10   & 57.3 & 9  & 9.5  & 5 FCV \\
UnpairedVSN~\cite{Rochan_2019_CVPR}  & 47.5          & 7      & 55.6           & 10      & 8.5 &  5 Rand \\
PCDL~\cite{8924889}    & 42.7  & 9   & 58.4 & 6  & 7.5  & 1 Rand \\
ACGAN~\cite{He:2019:UVS:3343031.3351056} & 46.0	       & 8      & 58.5           & 5      & 6.5  & 5 FCV \\
SUM-GAN-sl~\cite{Apostolidis:2019:SLA:3347449.3357482} & 47.8          & 6      & 58.4           & 6      & 6  &  5 Rand \\
SUM-GAN-AAE~\cite{10.1007/978-3-030-37731-1_40}    & 48.9  & 5   & 58.3 & 7  & 6 & 5 Rand \\
SUM-GDA$_{unsup}$~\cite{LI2021107677}  & 50.0 & 4      & 59.6           & 2      &  3 & 5 FCV \\
CSNet+GL+RPE~\cite{10.1007/978-3-030-58595-2_11}  & 50.2 & 3      & 59.1           & 3      &  3 & 5 FCV \\
CSNet~\cite{jung2019discriminative}  & \textbf{51.3} & 1      & 58.8           & 4      &  2.5 & 5 FCV \\
AC-SUM-GAN~\cite{9259058}  & 50.8 & 2      & \textbf{60.6}           & 1      &  \textbf{1.5} & 5 Rand \\
\hline
\end{tabular}}
\end{center}
\caption{Comparison (F1: F-Score (\%)) of unsupervised video summarization approaches on SumMe and TVSum.}
\label{tab:unsup_comp}
\end{table}

In Table \ref{tab:gt_comp} we show the performance of video summarization methods that have been evaluated with a variation of the established protocol, i.e., by comparing each generated summary with a single ground-truth summary per video (see Section \ref{subsubsec:eval_prot}, $3^{rd}$ paragraph). As before, this table reports the performance of a random summarizer according to this evaluation approach, provides details about the number and type of data splits that were used for the evaluation (see the rightmost column), and lists the algorithms according to their average ranking on both datasets (see the $4^{th}$ column ``Avg Rnk''). Our remarks on the reported data are as follows:
\begin{itemize}
    \item Only a limited number of video summarization works rely on the use of the single ground-truth summary for evaluating the summarization performance.
    \item Among the supervised methods, Ptr-Net is the top-performing one in both datasets. However, from the reportings in the relevant paper, it is not clear how many different randomly-created data splits were used for evaluation.
    \item Concerning the unsupervised approaches (marked with an asterisk) the SUM-GAN method and its extension that uses the Determinantal Point Process to increase the diversity of the summary content (called SUM-GAN$_{dpp}$) perform worse than a random summarizer. However, three newer extensions of this general approach that were evaluated also with this protocol (i.e., in addition to assessments made using the established evaluation approach), namely the SUM-GAN-sl, SUM-GAN-AAE and AC-SUM-GAN methods, exhibit very good performance that even surpasses the performance of the (few) supervised approaches listed in this table.
\end{itemize} 

\begin{table}
\begin{center}
\resizebox{\columnwidth}{!}{%
\begin{tabular}{|l|ll|ll|l|c|}
\hline
 & \multicolumn{2}{c|}{SumMe} & \multicolumn{2}{c|}{TVSum} & Avg & Data\\
\cline{2-5}
 & F1 & Rnk & F1 & Rnk  &  Rnk & splits\\ \hline
Random Summary & 40.2 & 8  & 54.4 & 9 & 8.5 & $-$  \\ \hline
$\star$SUM-GAN~\cite{Mahasseni2017UnsupervisedVS}   & 38.7 & 10 & 50.8 & 11 & 10.5 & 5 Rand \\
$\star$SUM-GAN$_{dpp}$~\cite{Mahasseni2017UnsupervisedVS}   & 39.1 & 9 & 51.7 & 10 & 9.5 & 5 Rand \\ 
SUM-GAN$_{sup}$~\cite{Mahasseni2017UnsupervisedVS} & 41.7 & 7  & 56.3 & 8  & 7.5 & 5 Rand\\
$\star$Cycle-SUM~\cite{Yuan2019CycleSUMCA} & 41.9 & 6  & 57.6 &  7 & 6.5 & 5 Rand\\
DTR-GAN~\cite{Zhang:2019:DDT:3321408.3322622}    & $-$ & $-$  & 61.3 & 6  & 6 & 1 Rand \\ 
Ptr-Net~\cite{DBLP:conf/wacv/FuTC19}    & 46.2 & 5  & 63.6 & 4  & 4.5 &  $-$    \\
$\star$SUM-$Ind_{LU}$~\cite{DBLP:journals/mta/YalinizI21}  & 51.4 & 3      & 61.5           & 5      &  4 & 5 FCV \\ 
$\star$SUM-GAN-sl~\cite{Apostolidis:2019:SLA:3347449.3357482}    & 46.8 & 4  & \textbf{65.3} & 1  & 2.5 &  5 Rand   \\
$\star$SUM-GAN-AAE~\cite{10.1007/978-3-030-37731-1_40}    & 56.9 & 2  & 63.9 & 3  & 2.5 &  5 Rand    \\
$\star$AC-SUM-GAN~\cite{9259058}    & \textbf{60.7} & 1  & 64.8 & 2  & \textbf{1.5} &  5 Rand    \\ \hline
\end{tabular}}
\end{center}
\caption{Comparison (F1: F-Score (\%)) of video summarization approaches on SumMe and TVSum, using a single ground-truth summary for each video. Unsupervised methods marked with $\star$.}
\label{tab:gt_comp}
\end{table}

Having discussed the results reported in each of the Tables \ref{tab:sup_comp}, \ref{tab:unsup_comp} and \ref{tab:gt_comp} alone, at this point we extend our observations by some additional remarks. The F-Score values reported in Tables \ref{tab:sup_comp} and \ref{tab:unsup_comp} show that the use of a supervision signal (associated with the ground-truth data) to train a method originally designed as an unsupervised one (e.g., DR-DSN, PCDL, ACGAN, CSNet) does not lead to considerably improved performance. Hence, focusing on purely supervised or unsupervised methods and trying to explore their learning capacity seems to be a more effective approach. Furthermore, purely unsupervised methods can be competitive to supervised ones (e.g., AC-SUM-GAN and CSNet), and thus additional future efforts towards the improvement of such algorithms are definitely in the right direction. 

With regards to the evaluation of video summarization algorithms, the rightmost column in all these three tables clearly indicates a lack of consistency. Most algorithms evaluate the summarization performance on $1$, $5$, or $10$ randomly-created data splits, while in some papers this information is completely missing (we denote this case by ``M Rand'' in these tables). Moreover, randomly-created data splits may exhibit some overlap among them, a case that differs from the 5-fold cross validation (see ``5 FCV'' in the tables) approach that is adopted by fewer methods. This observed diversity in the implementation of the employed evaluation protocol, along with the use of different randomly-created splits in most papers (see relevant considerations in \cite{10.1145/3394171.3413632}), unfortunately do not allow for a perfectly accurate performance comparison between the different summarization algorithms.

\begin{figure}[t]
\begin{center}
   \frame{\includegraphics[width=\columnwidth]{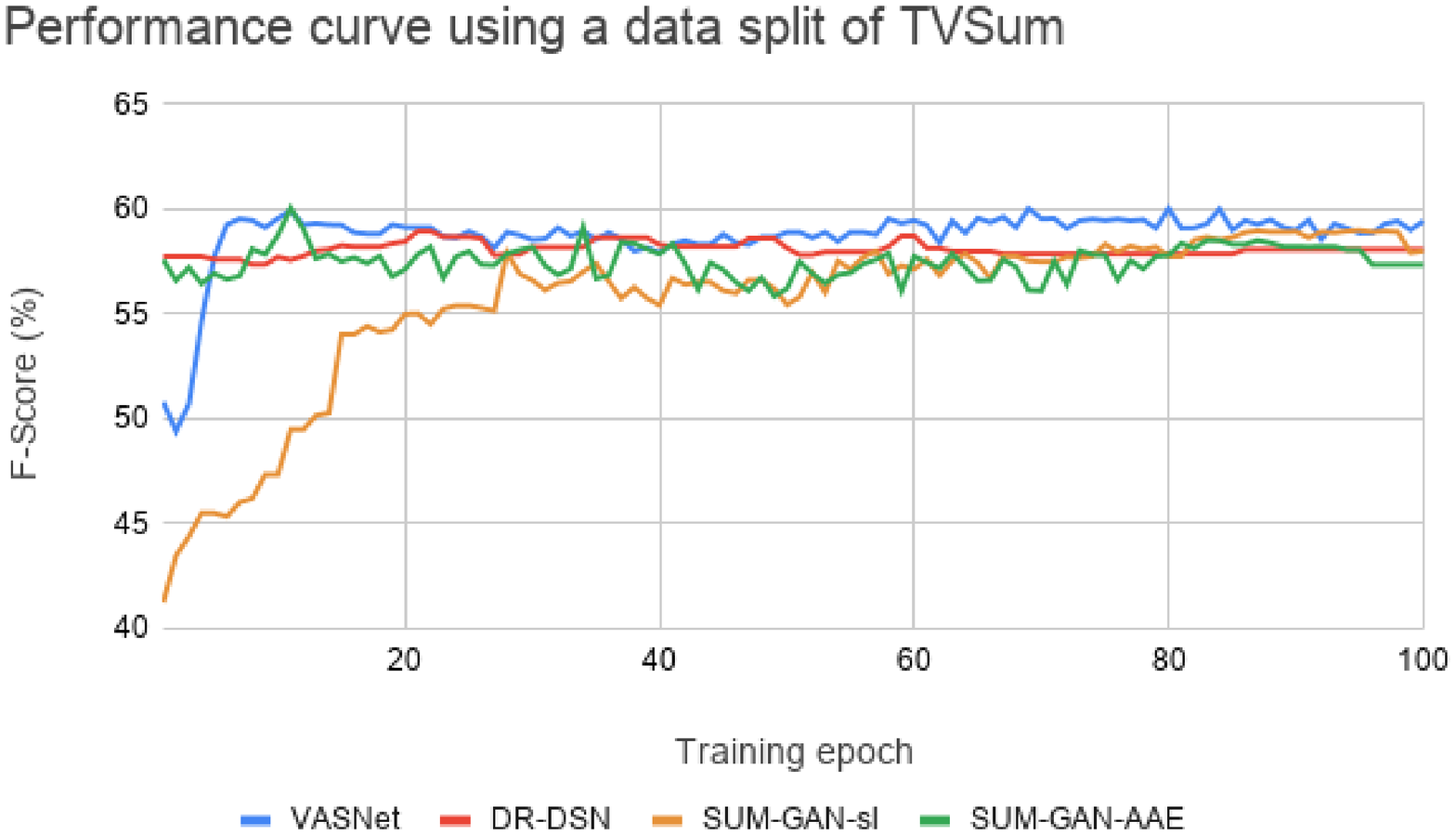}}
\end{center}
   \caption{Performance of four summarization methods on a set of test videos of the TVSum dataset as the training proceeds.}
\label{fig:performance_curve}
\end{figure}

Last but not least, another open issue of the relevant bibliography relates to a lack of information with respect to the applied approach for terminating the training process and selecting the trained model. Concerning the (weakly-) supervised methods, a few of them (e.g., \cite{10.1007/978-3-319-46478-7_47, 10.1007/978-3-030-01258-8_22, 8659061, Feng:2018:EVS:3240508.3240651, Wang:2019:SMN:3343031.3350992, Zhou2018VideoSB, yuan2019spatiotemporal}) explicitly state that model selection relies on the summarization performance on a validation set. One of them \cite{8101557} terminates the training process based on a pre-defined condition (training terminates if the average training loss difference between two consecutive epochs is less than a threshold that relies on the initial loss value). Most of them (e.g., \cite{Zhao:2017:HRN:3123266.3123328, ZhaoHSARNNHS, Zhang:2019:DDT:3321408.3322622, liu2019learning, DBLP:conf/wacv/FuTC19, Wei2018VideoSV, chu2019spatiotemporal, 8601376, 8924889, 9037206}) do not provide information regarding the use of a validation set or the application of a criterion for terminating the training process. Regarding the unsupervised approaches, one of them that relies on reinforcement learning \cite{Zhou2018DeepRL} states that training terminates upon a condition that relates to the received reward; i.e., training stops after reaching a maximum number of epochs (60 epochs), while early stopping is executed when the received reward stops to increase for a particular time period (10 epochs). Another recent approach \cite{9259058} that integrates an Actor-Critic model into a GAN and uses the discriminator's feedback as a reward signal, selects a well-trained model based on a criterion that maximizes the overall received reward and minimizes the Actor's loss. A couple of methods \cite{gonuguntla2019enhanced, jung2019discriminative} end the training process based on a maximum number of training epochs and then select the last trained model. Once again, most works (e.g., \cite{Mahasseni2017UnsupervisedVS, Rochan_2019_CVPR, He:2019:UVS:3343031.3351056, ZHANG2018, Yuan2019CycleSUMCA, Apostolidis:2019:SLA:3347449.3357482, 10.1007/978-3-030-37731-1_40, 10.1007/978-3-030-58595-2_11}) do not provide details about the use of a termination criterion. Nevertheless, experimentation with a few methods with publicly-available implementations (i.e., \cite{10.1007/978-3-030-21074-8_4, Zhou2018DeepRL, Apostolidis:2019:SLA:3347449.3357482, 10.1007/978-3-030-37731-1_40}) showed that the learning/performance curve can exhibit fluctuations; in Fig. \ref{fig:performance_curve}, the learning/performance curve of the examined supervised (VASNet) and unsupervised (DR-DSN, SUM-GAN-sl, SUM-GAN-AAE) algorithms shows noticeable fluctuations even after a good number of training epochs. This fluctuation denotes that the networks in general are able to develop knowledge about the task (sometimes even in the very early training epochs), but the selection of the best-trained model at the end is not always straightforward.  

\subsection{Qualitative Comparisons and Demos}

\begin{figure*}[]
\begin{center}
   \includegraphics[width=0.95\textwidth]{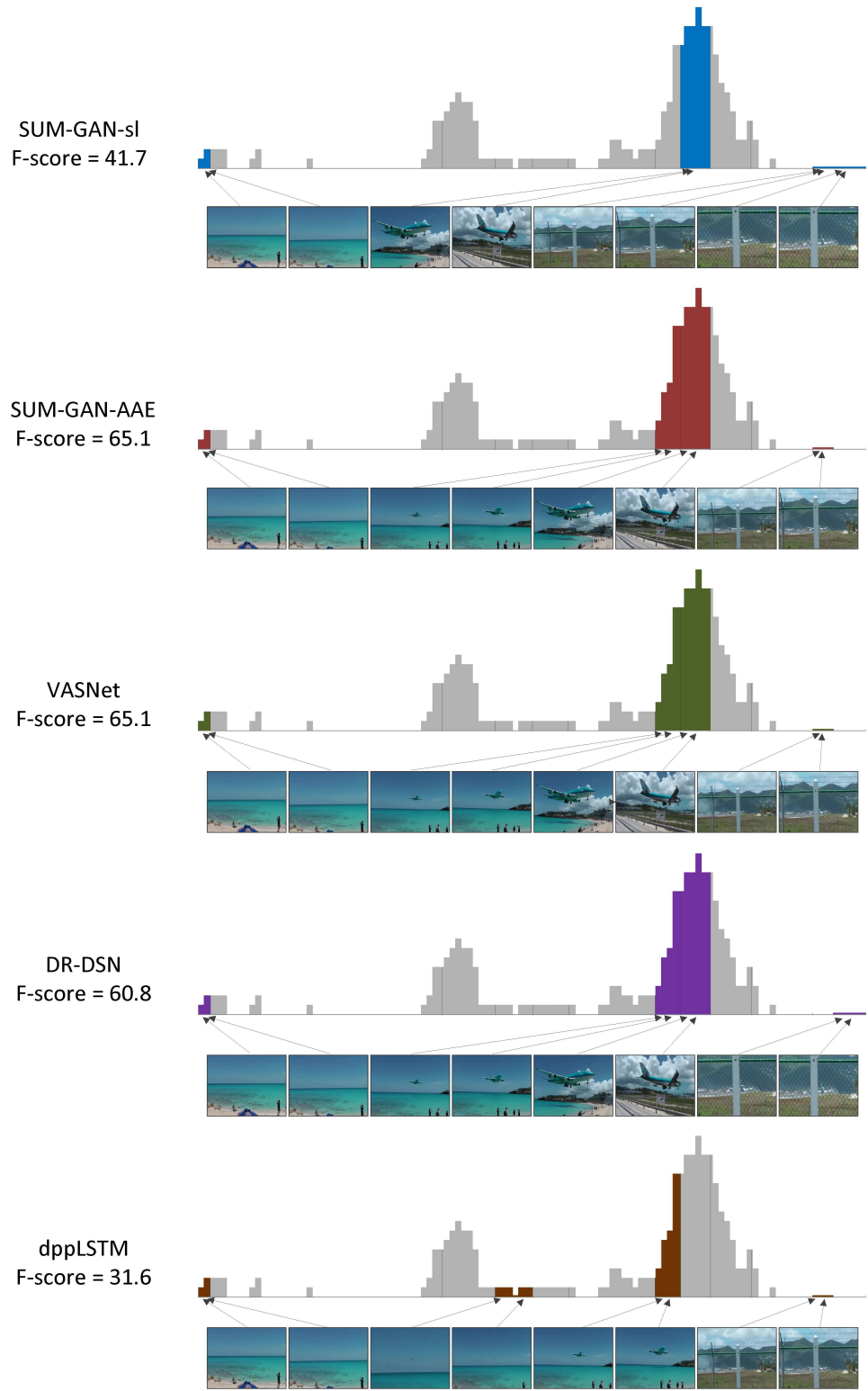}
\end{center}
   \caption{Overview of video $\#19$ of the SumMe dataset and the produced summaries by five summarization algorithms with publicly-available implementations.}
\label{fig:example_summaries}
\end{figure*}

In addition to the numerical comparisons discussed above, and with a view to gaining an intuitive understanding of the summarization results, in the sequel we present the summaries produced for video $\#19$ of SumMe (``St Maarten Landing'') by five publicly-available summarization methods. The frame sequence at the top of Fig.~\ref{fig:example_summaries} represents the flow of the story depicted in the video. Below this brief overview, for each considered method there is a diagram that shows the selected shots of the video and a visualization of the most important shots by a representative key-frame for each. Specifically, the gray bars represent the average human-annotated importance scores, the black vertical lines show the boundaries of each shot and the colored bars are the shots that each method has selected for inclusion in the summary. We observe that SUM-GAN-AAE and VASNet manage to select the shots with the highest importance, and for this they also achieve the highest F-score. Additionally, the key-frames they select are diverse and give a good overview of the plane's landing. DR-DSN selects almost the same shots as the two aforementioned methods, leading to a bit lower performance. SUM-GAN-sl focuses a bit less on the main event of the video; and dppLSTM loses the point of the video and selects many frames that show only the background without the plane, leading to a poor F-score.

Furthermore, to get an idea of how such summarization methods work in practice, one can experiment with tools such as the ``On-line Video Summarization Service''\footnote{http://multimedia2.iti.gr/videosummarization/service/start.html} of \cite{10.1145/3391614.3399391}, which integrates an adaptation of the SUM-GAN-AAE algorithm \cite{10.1007/978-3-030-37731-1_40}. This tool enables the creation of multiple summaries for a given video, which are tailored to the needs of different target distribution channels (i.e., different video sharing / social networking platorms).

\section{Future directions}
\label{future_directions}
Given the current state of the art in automated video summarization, we argue that future work in this field should primarily target the development of deep learning methods that can be trained effectively without the need for large collections of human-annotated ground-truth data. In this way, the research community will be able to tackle issues associated with the limited amount of annotated data, and to significantly diminish (or even completely eliminate) the need for laborious and time-demanding data annotation tasks. To this direction, the research community should put efforts towards the design and development of deep learning architectures that can be trained in a fully-unsupervised or in a semi/weakly-supervised manner.

With respect to the development of unsupervised video summarization methods, given the fact that most of the existing approaches try to increase the representativeness of the generated summary with the help of summary-to-video reconstruction mechanisms, future work could target the advancement of such methods by integrating mechanisms that force the outcome of the summarization process to meet additional criteria about the content of the generated summary, such as its visual diversity (that was considered in \cite{10.5555/2481023,LI2021107677,Rochan_2019_CVPR,Zhou2018DeepRL}) and its uniformity (that was examined in \cite{DBLP:journals/mta/YalinizI21}). On a similar basis, efforts could be put towards the extension of existing deep learning architectures that combine the merits of adversarial and reinforcement learning \cite{9259058}, by utilizing a Soft Actor-Critic \cite{Haarnoja2018SoftAO} that is capable of further discovering the action space via automatically defining a suitable value for the entropy regularization factor, and by introducing additional rewards that relate to the additional summarization criteria, such as the aforementioned ones.

With regards to the development of semi- or weakly-supervised approaches, the goal would be to investigate ways to intervene in the summary production process in order to force the outcome (i.e., a video summary) to be aligned with user-specified rules. One approach in this direction is the generation of a summary according to a set of textual queries that indicate the desired summary content (as in \cite{Sharghi2016QueryFocusedEV,Vasudevan:2017:QVS:3123266.3123297,DBLP:conf/bmvc/ZhangKLTX18,app9040750,10.1145/3372278.3390695}). Another, more aspiring approach would be the use of an on-line interaction channel between the user/editor and the trainable summarizer, in combination with active learning algorithms that allow to incorporate the user's/editor's feedback with respect to the generated summary (as in \cite{2491}). Finally, the possibility of adapting Graph Signal Processing approaches \cite{8347162}, which have already been applied with success to data sampling \cite{9244650} and image/video analysis tasks \cite{8334407,9288631}, for introducing such external supervision could be examined. The development of effective semi- or weakly-supervised summarization approaches will allow to better meet the needs of specific summarization scenarios and application domains. For example, such developments are often important for the practical application of summarization technologies in the News/Media Industry, where complete automation that diminishes editorial control over the generated summaries is not always preferred.

Concerning the training of unsupervised video summarization methods, we show that most of these methods rely on the adversarial training of GANs. However, open questions with respect to the training of such architectures, such as sufficient convergence conditions and mode collapse, still remain. So, another promising research direction could be to investigate ways to improve the training process. For this, one strategy could be the use of augmented training data (that do not require human annotation) in combination with curriculum learning approaches. Such approaches have already been examined for improving the training of GANs (see \cite{doan2019on-line,8954410,9093408}) in applications other than video summarization. We argue that transferring the gained knowledge from these works to the video summarization domain would contribute to advancing the effectiveness of unsupervised GAN-based summarization approaches. Regarding the training of semi- or weakly-supervised video summarization methods, besides the use of an on-line interaction channel between the user/editor and the trainable summarizer that was discussed in the previous paragraph, supervision could also relate to the use of an adequately-large set of unpaired data (i.e., raw videos and video summaries with no correspondence between them) from a particular summarization domain or application scenario. Taking inspiration from the method in \cite{Rochan_2019_CVPR}, we believe that such a data-driven weak-supervision approach would eliminate the need for fine-grained supervision signals (i.e., human-generated ground-truth annotations for the collection of the raw videos) or hand-crafted functions that model the domain rules (which in most cases are really hard to obtain), and would allow a deep learning architecture to automatically learn a mapping function between the raw videos and the summaries of the targeted domain.	
  
Another future research objective involves efforts to overcome the identified weaknesses of using RNNs for video summarization that were discussed in e.g., \cite{10.1007/978-3-030-21074-8_4,9037206,LI2021107677,DBLP:journals/mta/YalinizI21} and mainly relate to the computationally-demanding and hard-to-parallelize training process, as well as to the limited memory capacity of these networks. For this, future work could examine the use of Independently Recurrent Neural Networks \cite{Li2018IndependentlyRN} that were shown to alleviate the drawbacks of LSTMs with respect to decaying, vanishing and exploding gradients \cite{DBLP:journals/mta/YalinizI21}, in combination with high-capacity memory networks, such as the ones used in \cite{Feng:2018:EVS:3240508.3240651,Wang:2019:SMN:3343031.3350992}. Alternatively, future work could build on existing approaches \cite{10.1007/978-3-030-21074-8_4,LI2021107677,8601376,10.1007/978-3-030-58595-2_11} and develop more advanced attention mechanisms that encode the relative position of video frames and model their temporal dependencies according to different granularities (e.g., considering the entire frame sequence, or also focusing on smaller parts of it). Such methods would be particularly suited for summarizing long videos (e.g., movies). Finally, with respect to video content representation, the above proposed research directions could also involve the use of network architectures that model the spatiotemporal structure of the video, such as 3D-CNNs and convolutional LSTMs.
	
With respect to the utilized data modality for learning a summarizer, currently the focus is on the visual modality. Nevertheless, the audio modality of the video could be a rich source of information as well. For example, the audio content could help to automatically identify the most thrilling parts of a movie that should appear in a movie trailer. Moreover, the temporal segmentation of the video based also on the audio stream could allow the production of summaries that offer a more natural story narration compared to the generated summaries based on approaches that rely solely on the visual stream. We argue that deep learning architectures that have been utilized to model frames' dependencies based on their visual content, could be examined also for analyzing the audio modality. Following, the extracted representations from these two modalities could be fused according to different strategies (e.g., after exploring the latent consistency between them), to better indicate the most suitable parts for inclusion in the video summary.

Finally, besides the aforementioned research directions that relate to the development and training of deep-learning-based architectures for video summarization, we strongly believe that efforts should be put towards the definition of better evaluation protocols to allow accurate comparison of the developed methods in the future. The discussions in \cite{otani2018vsumeval} and \cite{10.1145/3394171.3413632} showed that the existing protocols have some imperfections that affect the reliability of performance comparisons. To eliminate the impact of the choices made when evaluating a summarization algorithm (that e.g., relate to the split of the utilized data or the number of different runs), the relevant community should consider all the different parameters of the evaluation pipeline and precisely define a protocol that leaves no questions about the experimental outcomes of a summarization work. Then, the adoption of this protocol by the relevant community will enable fair and accurate performance comparisons.

\section{Conclusions}
\label{sec:conclusions}
In this work we provided a systematic review of the deep-learning-based video summarization landscape. This review allowed to discuss how the summarization technology has evolved over the last years and what is the potential for the future, as well as to raise awareness to the relevant community with respect to promising future directions and open issues. The main conclusions of this study are outlined in the following paragraphs.  

Concerning the summarization performance, the best-performing supervised methods thus far learn frames' importance by modeling the variable-range temporal dependency among video frames/fragments with the help of Recurrent Neural Networks and tailored attention mechanisms. The extension of the memorization capacity of LSTMs by using memory networks has shown promising results and should be further investigated. In the direction of unsupervised video summarization, the use of Generative Adversarial Networks for learning how to build a representative video summary seems to be the most promising approach. Such networks have been integrated in summarization architectures and used in combination with attention mechanisms or Actor-Critic models, showing a summarization performance that is comparable to the performance of state-of-the-art supervised approaches. Given the objective difficulty to create large-scale datasets with human annotations for training summarization models in a supervised way, further research effort should be put on the development of fully-unsupervised or semi/weakly-supervised video summarization methods that eliminate or reduce to a large extent the need for such data, and facilitate adaptation to the summarization requirements of different domains and application scenarios.

Regarding the evaluation of video summarization algorithms, there is some diversity among the used evaluation protocols in the bibliography, that is associated to the way that the used data are being divided for training and testing purposes, the number of the conducted experiments using different randomly-created splits of the data, and the used data splits; concerning the latter, a recent work \cite{10.1145/3394171.3413632} showed that different randomly-created data splits of the SumMe and TVSum datasets are characterized by considerably different levels of difficulty. All the above raise concerns regarding the accuracy of performance comparisons that rely on the results reported in the different papers. Moreover, there is lack of information in the reportings of several summarization works, with respect to the applied process for terminating the training process and selecting the trained model. Hence, the relevant community should be aware of these issues and take the necessary actions to increase the reproducibility of the results reported for each newly-proposed method.

Last but not least, this work indicated several research directions towards further advancing the performance of video summarization algorithms. Besides these proposals for future scientific work, we believe that further efforts should be put towards the practical use of summarization algorithms, by integrating such technologies into tools that support the needs of modern media organizations for time-efficient video content adaptation and re-use. 

\ifCLASSOPTIONcaptionsoff
  \newpage
\fi


\bibliographystyle{IEEEtran}
\bibliography{survey}


\begin{IEEEbiography}[{\includegraphics[width=1in,height=1.25in,clip,keepaspectratio]{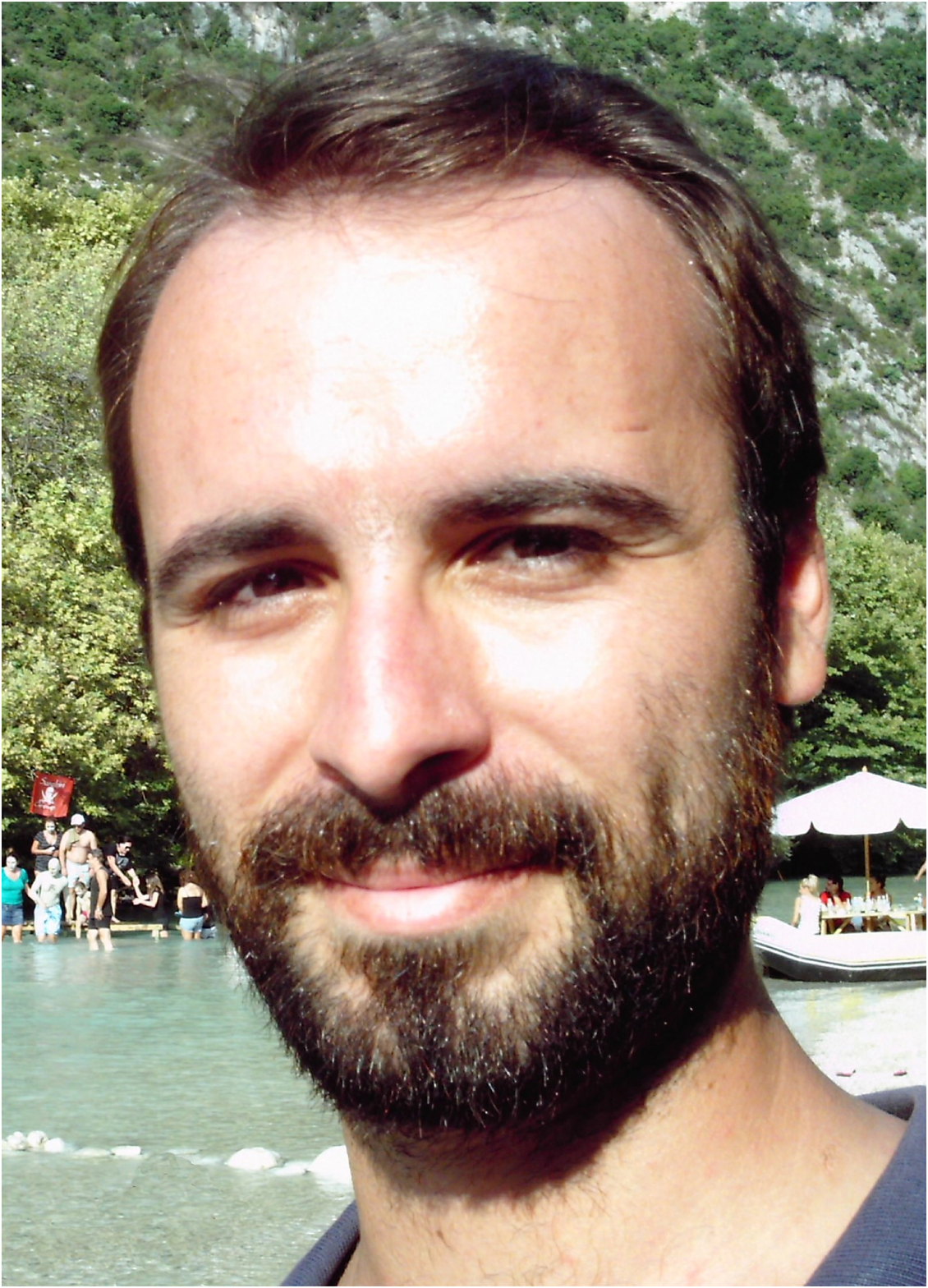}}]{Evlampios Apostolidis}
received the Diploma degree in Electrical and Computer Engineering from the Aristotle University of Thessaloniki, Greece, in 2007. His diploma thesis was on methods for digital watermarking of 3D TV content. Following, he received a M.Sc. degree in Information Systems from the University of Macedonia, Thessaloniki, Greece, in 2011. For his Dissertation he studied techniques for indexing multi-dimensional data. Since January 2012 he is a Research Assistant at the Centre for Research and Technology Hellas - Information Technologies Institute. Since May 2018 he is also a PhD student at the Queen Mary University of London - School of Electronic Engineering and Computer Science. He has co-authored 3 journal papers, 4 book chapters and more than 25 conference papers. His research interests lie in the areas of video analysis and understanding, with a particular focus on methods for video segmentation and summarization.
\end{IEEEbiography}

\begin{IEEEbiography}[{\includegraphics[width=1in,height=1.25in,clip,keepaspectratio]{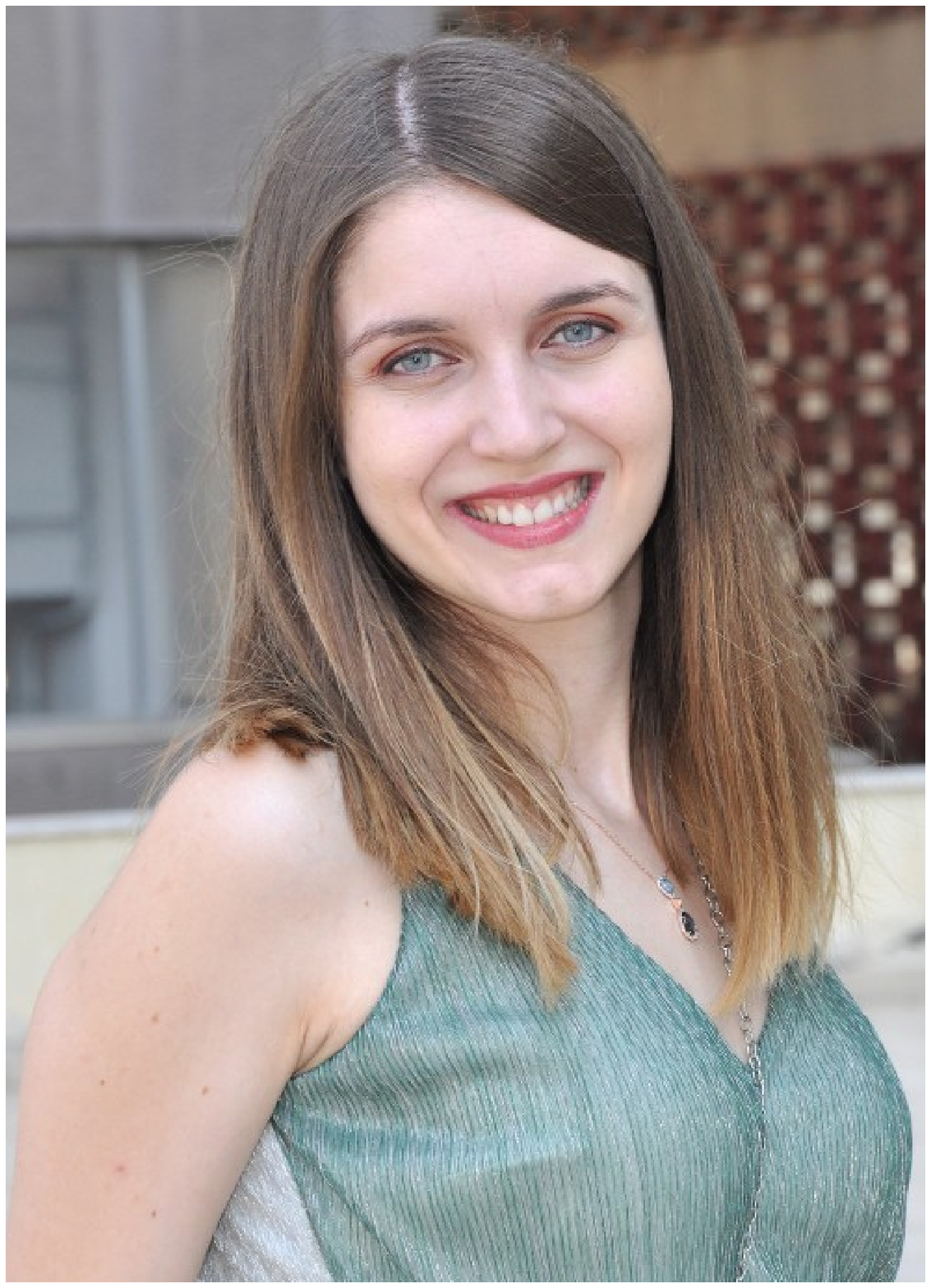}}]{Eleni Adamantidou}
received her Diploma degree in Electrical and Computer Engineering from the Aristotle University of Thessaloniki, Greece, in 2019. She graduated in the top $10\%$ of her class (grade 9.01/10). Since March 2019 she works as a Research Assistant at the Centre for Research and Technology Hellas - Information Technologies Institute. She has co-authored 1 journal and 5 conference papers in the field of video summarization. She is particularly interested in deep learning methods for video analysis and summarization, and natural language processing.
\end{IEEEbiography}

\begin{IEEEbiography}[{\includegraphics[width=1in,height=1.25in,clip,keepaspectratio]{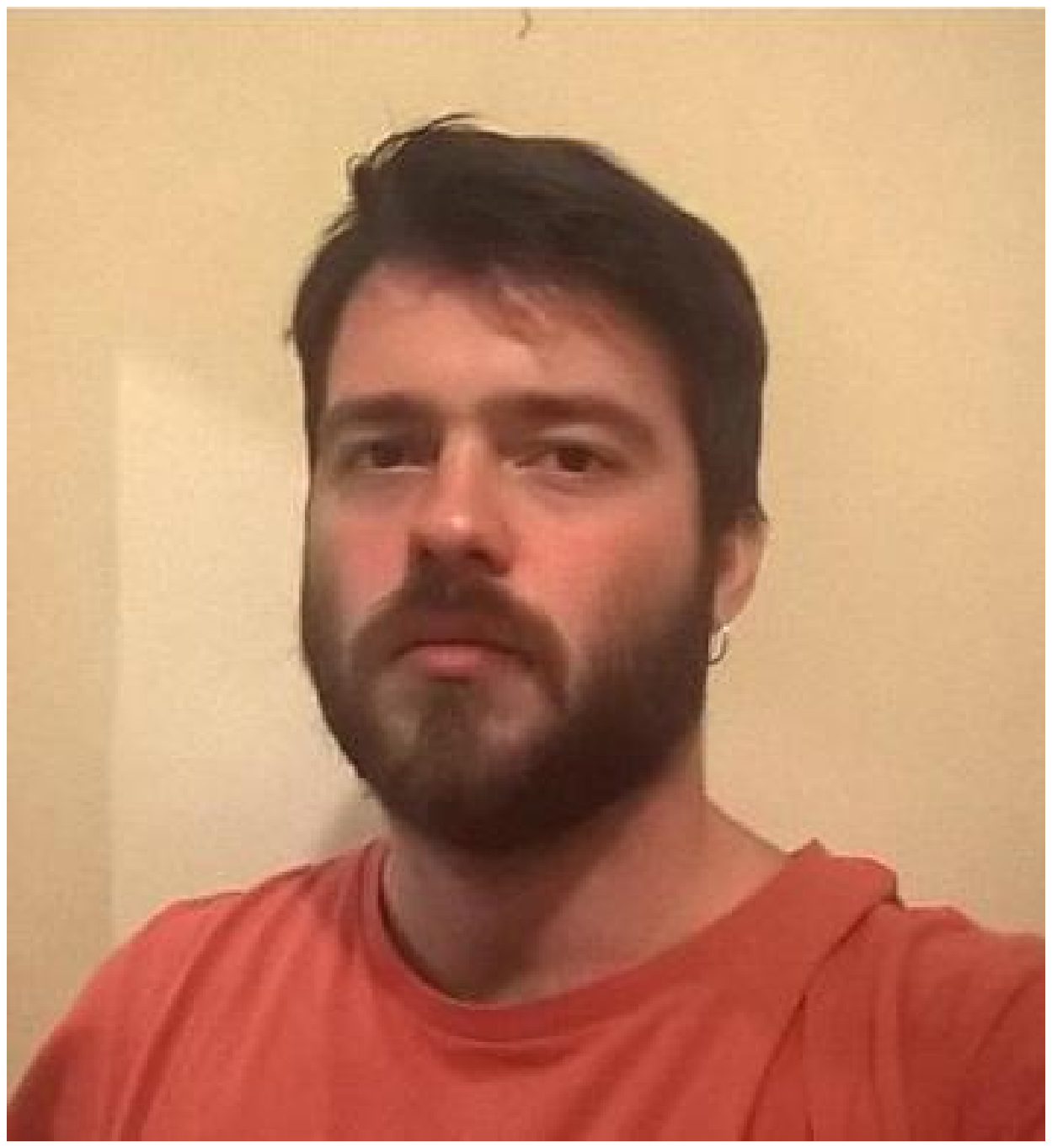}}]{Alexandros I. Metsai}
received the Diploma degree in Electrical and Computer Engineering from the Aristotle University of Thessaloniki, Greece, in 2017. For the needs of his diploma thesis, he developed a robotic agent capable of learning objects shown by a human through voice commands and hand gestures. Since September 2018 he works as a Research Assistant at the Centre for Research and Technology Hellas - Information Technologies Institute. He has co-authored 1 journal and 4 conference papers in the field of video summarization and 1 book chapter in the field of video forensics. His research interests are in the area of deep learning for video analysis and summarization.
\end{IEEEbiography}

\begin{IEEEbiography}[{\includegraphics[width=1in,height=1.25in,clip,keepaspectratio]{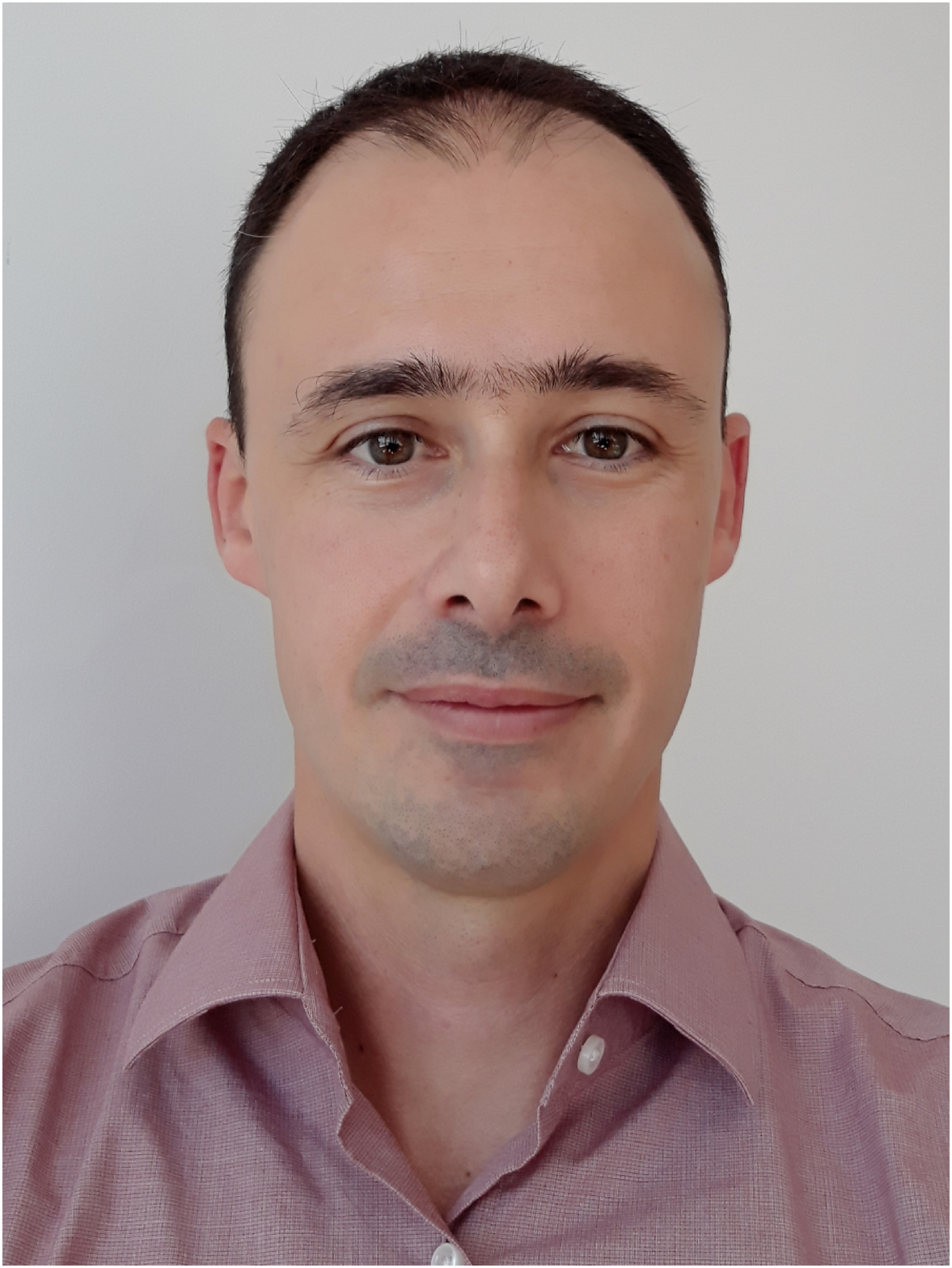}}]{Vasileios Mezaris}
received the B.Sc. and Ph.D. degrees in Electrical and Computer Engineering from Aristotle University of Thessaloniki, Greece, in 2001 and 2005, respectively. He is currently a Research Director with the Centre for Research and Technology Hellas - Information Technologies Institute. He has co-authored over 40 journal papers, 20 book chapters, 170 conference papers, and three patents. His research interests include multimedia understanding and artificial intelligence; in particular, image and video analysis and annotation, machine learning and deep learning for multimedia understanding and big data analytics, multimedia indexing and retrieval, and applications of multimedia understanding and artificial intelligence. He serves as a Senior Area Editor for IEEE SIGNAL PROCESSING LETTERS and as an Associate Editor for IEEE TRANSACTIONS ON MULTIMEDIA. He is a senior member of the IEEE.
\end{IEEEbiography}

\begin{IEEEbiography}[{\includegraphics[width=1in,height=1.25in,clip,keepaspectratio]{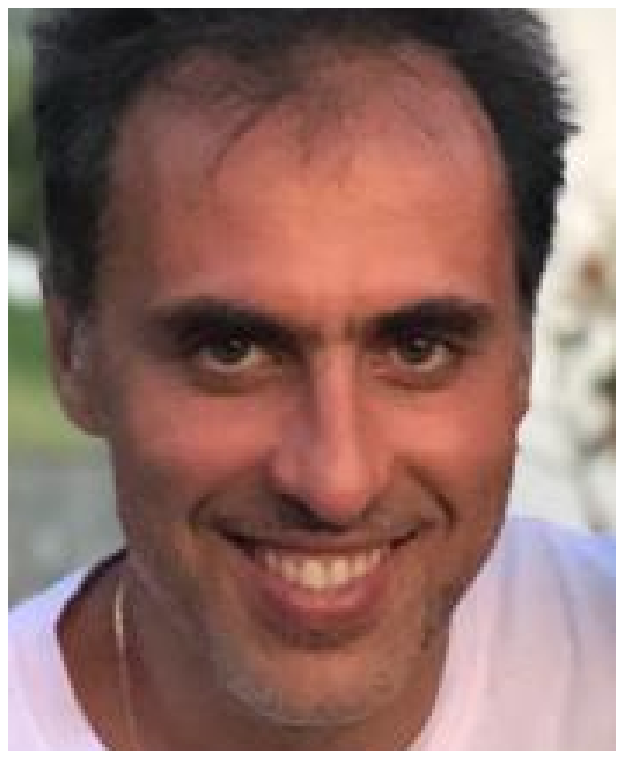}}]{Ioannis (Yiannis) Patras} 
is a Professor in Computer Vision and Human Sensing in the School of Electronic Engineering and Computer Science in the Queen Mary University of London. He is/has been in the organizing committee of Multimedia Modeling 2021, ICMR 2019, IEEE Image, Video and Multidimensional Signal Processing Symposium 2018, ACM Multimedia 2013, ICMR 2011, Face and Gesture Recognition 2008, BMVC 2009, and was the general chair of WIAMIS 2009. He is associate editor in the journal of Pattern Recognition, Computer Vision and Image Understanding and the Image and Vision Computing Journal, and Area chair in major Computer Vision conferences including, ICCV, ECCV, FG and BMVC. He has more than 200 publications in the most selective journals and conferences in the field of Computer Vision. He is a senior member of IEEE and a member of the Visual Signal Processing and Communications Technical Committee (VSPC) of CAS society. His research interests lie in the areas of Computer Vision and Human Sensing using Machine Learning methodologies -- this includes tracking and recognition of human actions and activity in multimedia, affect analysis and multimodal analysis of human behaviour.
\end{IEEEbiography}

\end{document}